\documentclass[11pt]{knowchange}
\usepackage{fix-cm}
\usepackage{helvet}
\usepackage{xcolor}
\usepackage{multirow}
\usepackage{colortbl}
\usepackage{tabularx}
\usepackage{tcolorbox}
\usepackage{enumitem}
\usepackage{pifont}
\usepackage{wrapfig}
\usepackage{url}
\usepackage{xspace}
\usepackage{amsmath,amssymb,amsfonts,bm,mathtools}
\usepackage{adjustbox}
\usepackage{array}
\usepackage{float}
\usepackage{booktabs}
\usepackage{siunitx}
\usepackage{pdflscape}
\usepackage{placeins}
\usepackage{tikz}
\usepackage{pgfplots}
\pgfplotsset{compat=1.18}
\usepackage{algorithm}
\usepackage{algpseudocode}
\usepackage{listings}
\usepackage{subcaption}
\usepackage{dsfont}
\usepackage{cleveref}
\hypersetup{colorlinks=true}

\newcolumntype{L}[1]{>{\raggedright\let\newline\\\arraybackslash\hspace{0pt}}m{#1}}
\newcolumntype{R}[1]{>{\raggedleft\let\newline\\\arraybackslash\hspace{0pt}}m{#1}}

\newcommand{\ignore}[1]{}

\makeatletter
\DeclareRobustCommand\onedot{\futurelet\@let@token\@onedot}
\def\@onedot{\ifx\@let@token.\else.\null\fi\xspace}

\def\eg{e.g\onedot}

\makeatother

\definecolor{MyBlue}{rgb}{0.46, 0.50, 0.61}
\definecolor{MyDarkBlue}{rgb}{0,0.08,0.8}
\definecolor{MyDarkGreen}{RGB}{45,155,45}
\definecolor{MyDarkRed}{rgb}{0.8,0.02,0.02}
\definecolor{MyOrange}{rgb}{1.0, 0.4, 0.2}
\definecolor{MyPurple}{RGB}{111,0,255}
\definecolor{MyRed}{rgb}{0.8,0.0,0.0}
\definecolor{MyGold}{rgb}{0.75,0.6,0.12}
\definecolor{MyDarkgray}{rgb}{0.66, 0.66, 0.66}
\definecolor{MyBrown}{rgb}{0.65, 0.16, 0.16}
\definecolor{MyMutedRose}{rgb}{0.58, 0.29, 0.35}
\definecolor{JiayuanColor}{rgb}{0.60,0.43,0.48}
\definecolor{erranColor}{rgb}{24, 40, 113}

\definecolor{citecolor}{HTML}{696FAD}

\newif\ifpropositionfirstitem
\propositionfirstitemtrue

\definecolor{bggray}{HTML}{F5F5F5}
\definecolor{pvdblue}{HTML}{DAE8FC}
\definecolor{RoseQuartzBg}{HTML}{F7CAC9}
\definecolor{RoseQuartz}{HTML}{F5A798}
\definecolor{Serenity}{HTML}{92A8D1}
\definecolor{OrangeRed}{rgb}{1.0, 0.27, 0.0}
\definecolor{RoyalBlue}{cmyk}{1, 0.50, 0, 0}
\definecolor{Turquoise}{HTML}{0F4C81}
\definecolor{mint}{rgb}{0.24, 0.71, 0.54}
\definecolor{green}{rgb}{0.0, 0.120, 0.0}

\newdimen\abovecrulesep
\newdimen\belowcrulesep
\makeatletter
\patchcmd{\@@@cmidrule}{\aboverulesep}{\abovecrulesep}{}{}
\patchcmd{\@xcmidrule}{\belowrulesep}{\belowcrulesep}{}{}
\makeatother

\definecolor{mybluetitle}{HTML}{4B527E}%

\definecolor{mygreen}{RGB}{0,150,0}
\definecolor{boxbackground}{HTML}{F0F7FF}%
\definecolor{boxborder}{HTML}{D0D9E5}%
\definecolor{accentblue}{HTML}{4A86E8}%
\definecolor{lightblue}{HTML}{EEF3FF}%
\definecolor{bordergray}{HTML}{CCCCCC}%
\definecolor{headerblue}{HTML}{2C5AA0}%

\definecolor{lavenderframe}{HTML}{E6E6FA}%
\definecolor{lighterlav}{HTML}{F5F5FF}%
\definecolor{codegray}{rgb}{0.5,0.5,0.5}%
\definecolor{codepurple}{HTML}{483D8B}%
\definecolor{backcolour}{HTML}{F5F5FF}%
\lstdefinestyle{mystyle}{
    backgroundcolor=\color{backcolour},
    commentstyle=\color{headerblue},
    keywordstyle=\color{codepurple},
    numberstyle=\tiny\color{codegray},
    stringstyle=\color{codepurple},
    basicstyle=\ttfamily\scriptsize,
    breakatwhitespace=false,
    breaklines=true,
    captionpos=b,
    keepspaces=true,
    frame=none,
    numbersep=5pt,
    showspaces=false,
    showstringspaces=false,
    showtabs=false,
    tabsize=2
}

\definecolor{jsonkey}{RGB}{44, 130, 201}%
\definecolor{jsonstring}{RGB}{255, 140, 0}%
\definecolor{jsonnumber}{RGB}{34, 139, 34}%

\lstdefinelanguage{json}{
    basicstyle=\ttfamily\small,
    numbers=left,
    numberstyle=\tiny\color{gray},
    stepnumber=1,
    numbersep=5pt,
    showstringspaces=false,
    breaklines=true,
    frame=none,
    backgroundcolor=\color{gray!5},
    literate=
     *{:}{{{\color{jsonkey}:}}}{1}
      {,}{{{\color{jsonkey},}}}{1}
      {"}{{{\color{jsonstring}"}}}{1}
      {[}{{{\color{jsonkey}[}}}{1}
      {]}{{{\color{jsonkey}]}}}{1}
      {0}{{{\color{jsonnumber}0}}}{1}
      {1}{{{\color{jsonnumber}1}}}{1}
      {2}{{{\color{jsonnumber}2}}}{1}
      {3}{{{\color{jsonnumber}3}}}{1}
      {4}{{{\color{jsonnumber}4}}}{1}
      {5}{{{\color{jsonnumber}5}}}{1}
      {6}{{{\color{jsonnumber}6}}}{1}
      {7}{{{\color{jsonnumber}7}}}{1}
      {8}{{{\color{jsonnumber}8}}}{1}
      {9}{{{\color{jsonnumber}9}}}{1}
}

\newtcblisting{jsonbox}{
  listing engine=listings,
  colback=gray!3!white,
  colframe=gray!75!black,
  boxrule=0.4mm,
  arc=2mm,
  outer arc=2mm,
  breakable,
  enhanced,
  listing only,
  listing options={language=json}
}

\newtcolorbox{promptbox}[2][]{%
    enhanced,
    breakable,
    boxsep=5pt,
    left=9pt,
    right=7pt,
    top=5pt,
    bottom=5pt,
    colback=boxbackground,
    colframe=boxborder,
    boxrule=0.5pt,
    arc=4pt,
    frame hidden,%
    borderline west={3pt}{0pt}{accentblue},
    shadow={0.5pt}{0.5pt}{1.5pt}{black!10},
    fontupper=\normalsize,
    title=#2,%
    colbacktitle=accentblue,%
    coltitle=white,%
    fonttitle={\fontsize{9}{11}\selectfont\bfseries},%
    attach boxed title to top left={yshift=-2.5mm, xshift=3.2mm},
    boxed title style={
        enhanced,
        left=3pt,
        right=3pt,
        top=1pt,%
        bottom=1pt,%
        boxsep=2pt,
        arc=3pt,
        boxrule=0pt,
        colback=accentblue,
    },
    #1%
}

\newtcolorbox{notitlepromptbox}[1][]{
    enhanced,
    breakable,
    boxsep=5pt,%
    left=9pt,%
    right=7pt,%
    top=5pt,%
    bottom=5pt,%
    colback=boxbackground,
    colframe=boxborder,
    boxrule=0.5pt,
    arc=4pt,%
    frame hidden,
    borderline west={3pt}{0pt}{accentblue},%
    shadow={0.5pt}{0.5pt}{1.5pt}{black!10},%
    notitle,
    fontupper=\normalsize,%
    #1
}

\newtcolorbox{onebox}[2][]{
    enhanced, 
    center title,
    left*=0pt, right*=0pt,
    boxsep=2pt, left=5pt, right=5pt,
    skin first=enhanced,
    skin middle=enhanced,
    skin last=enhanced,
    colframe = mybluetitle!90,
  colback  = mybluetitle!10,
    fonttitle=\bfseries\rmfamily\fontfamily{phv}\selectfont,
    title={\footnotesize\strut{#2}  \refstepcounter{subsubsection} \addcontentsline{toc}{subsubsection}{\string\numberline{\thesubsubsection}#2}
    },
    #1
    }

\title{Real-World Knowledge-Guided Change Data Synthesis for Remote Sensing}

\author[1,*]{Yaoyi Qi}
\author[1,*]{Xingxing Weng}
\author[1,\textdagger]{Chao Pang}
\author[1]{Yongkang Cui}
\author[1]{XiangYu Hao}
\author[1]{Xiaokang Zhang}
\author[2,3]{Guibo Zhu}
\author[1,4]{Gui-Song Xia}

\affiliation[1]{School of Artificial Intelligence, Wuhan University}
\affiliation[2]{Wuhan AI Research}
\affiliationbreak
\affiliation[3]{Institute of Automation, University of Chinese Academy of Sciences}
\affiliation[4]{Institute for Math \& AI, Wuhan}
\correspondence{pangchao@whu.edu.cn}
\contribution[*]{Equal contribution}

\abstract{Change data synthesis provides a cost-effective solution for expanding training data and improving the performance of change detection models. However, existing synthesis methods typically rely on handcrafted rules to simulate changes, where limited coverage of class transitions restricts the diversity of synthesized data, while predefined transition designs limit their flexibility in accommodating varied change types. In this work, we introduce KnowChange, a knowledge-guided change data synthesis framework that leverages pretrained vision-language models as knowledge sources to reason about plausible change locations and class transitions from pre-change scenes and desired change types. By integrating knowledge-guided change simulation with generalizable synthesis models, KnowChange enables flexible synthesis of diverse change types within a unified framework. Extensive experiments demonstrate that KnowChange-generated data consistently outperforms existing synthetic datasets in both synthetic-to-real transfer and synthetic data augmentation, despite being generated at a compact scale. Further analyses show that the knowledge-guided change simulation can be seamlessly integrated into existing synthesis pipelines and enhance the downstream utility of synthesized data.

\vspace{8pt}

~
\textbf{Web}: \url{https://knowchange.vercel.app/} \\~
\textbf{Code}: \url{https://github.com/LINGQI711/KnowChange} \\~

}

\begin{document}
\maketitle

\section{Introduction}
\label{sec:intro}
Change data synthesis aims to automatically generate bi-temporal images with pixel-level change masks, and optionally semantic masks for each timestamp to characterize class transitions between the two images. By reducing reliance on expensive manual annotation, change data synthesis provides a scalable solution for increasing training data diversity and has attracted growing attention in remote sensing~\cite{zheng2024changen2,zan2025open,liu2025generating}.

\begin{figure}[t]
    \centering
    \includegraphics[width=0.95\linewidth]{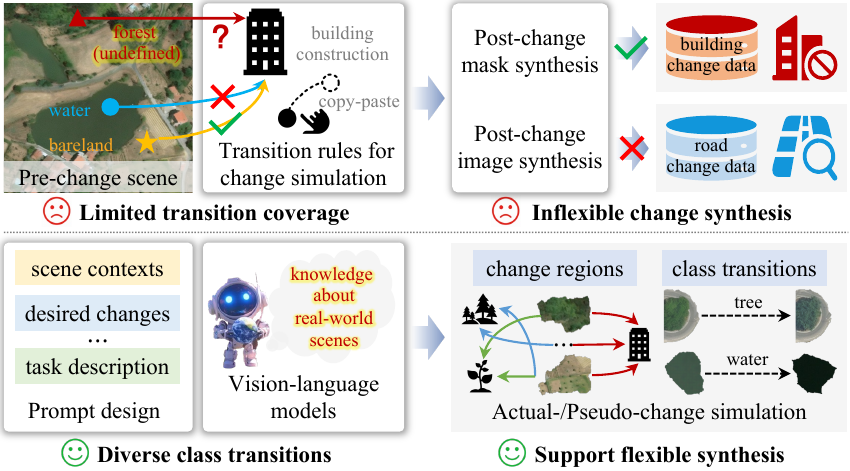}
    \caption{\textbf{Top}: Existing synthesis methods rely on handcrafted rules to simulate changes. Limited coverage of class transitions restricts the diversity of synthesized data, while predefined transition designs limit flexibility in accommodating varied change types. \textbf{Bottom}: KnowChange leverages pretrained vision-language models as knowledge sources to reason about plausible change regions and class transitions given pre-change scene contexts and user-specified change types, enabling flexible synthesis of diverse change data.}
    \label{fig:comparison-method}
\end{figure}

Existing change data synthesis methods~\cite{zheng2024changen2,benidir2025change,liu2025generating} typically start from single-temporal images with semantic masks and simulate future changes to obtain post-change semantic masks, which are then used to guide the generation of post-change images. In this pipeline, change simulation is crucial, as it determines where changes occur and what class transitions take place. Many methods implement this simulation through handcrafted rules that specify class transitions and guide region-level manipulations (\emph{e.g.}, copy-paste). Although rule-based change simulation has enabled the construction of large-scale synthetic datasets, it suffers from two fundamental limitations, as illustrated in Fig.~\ref{fig:comparison-method}.

First, handcrafted rules usually cover only a limited set of class transitions. Due to the large field of view and diverse land-cover categories in remote sensing images, real-world changes often involve a broader transition space than predefined rules can capture. For example, existing rules allow only a few land-cover categories, such as bareland, rangeland, and developed land, to transition into buildings, whereas real-world urban expansion can also involve transitions from forests or other land-cover categories into buildings. Such limited transition coverage restricts the diversity of synthesized change data, potentially constraining the performance of change detection models trained on synthetic data.

Second, desired change types vary across application scenarios. For example, urban development monitoring involves building construction or demolition, whereas transportation monitoring concerns road-related changes. However, rule-based change simulation relies on predefined transitions with fixed patterns, limiting its flexibility in accommodating new change types. Supporting new change types requires redesigning the transition rules and adapting the image synthesis models accordingly.

Existing change simulation largely relies on human knowledge about real-world changes encoded in handcrafted rules. However, manually enumerating and encoding such knowledge into rules for diverse and evolving change scenarios is inherently difficult. Recent vision-language models (VLMs) pretrained on large-scale vision-language corpora capture rich visual-semantic knowledge about real-world scenes, including object categories, their relationships, and scene contexts. This motivates us to leverage VLMs as a knowledge source to infer where changes can occur and what class transitions are plausible, enabling real-world knowledge-guided change simulation. 

\begin{figure*}[t]
    \centering
    \includegraphics[width=\textwidth]{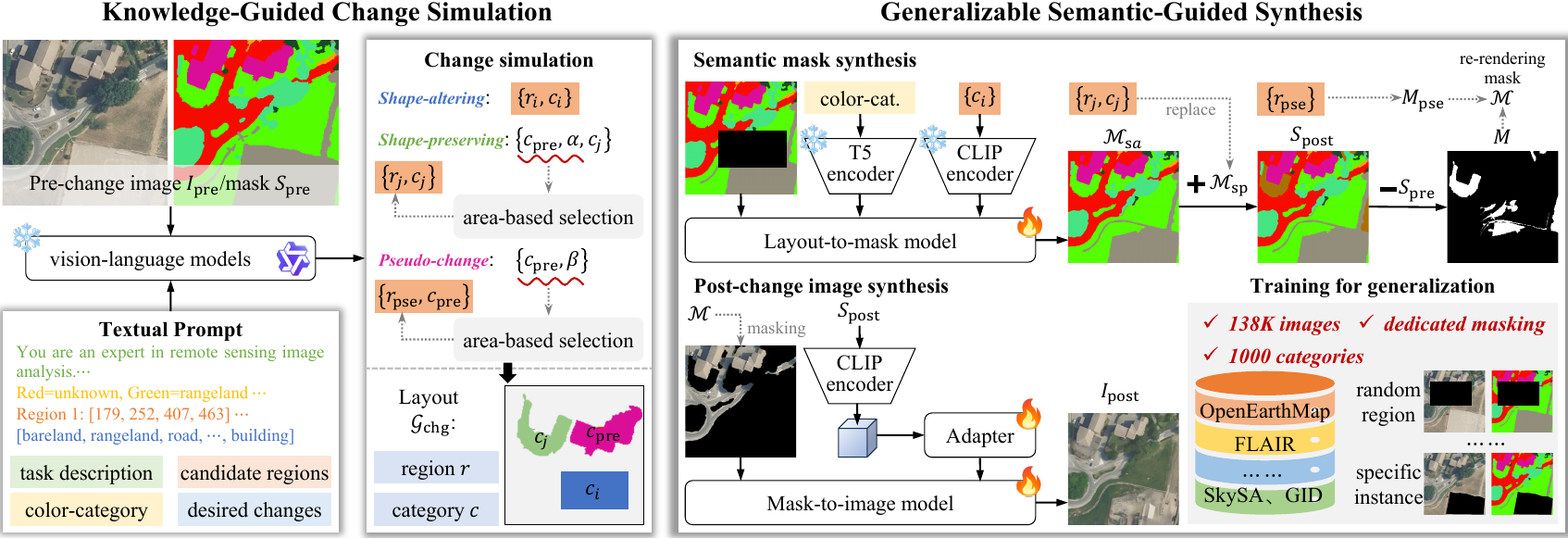}
\caption{
Overview of KnowChange, a knowledge-guided framework for change data synthesis.
Given a pre-change image $I_{\mathrm{pre}}$, its semantic mask $S_{\mathrm{pre}}$, and a textual prompt $P$, Knowledge-Guided Change Simulation reasons about actual and pseudo changes to construct the change layout $\mathcal{G}_{\mathrm{chg}}$.
Generalizable Semantic-Guided Synthesis generates $\mathcal{M}_{\mathrm{sa}}$, combines it with $\mathcal{M}_{\mathrm{sp}}$ to obtain $S_{\mathrm{post}}$ and $\mathcal{M}$, and uses them to guide the synthesis of the post-change image $I_{\mathrm{post}}$.
}
\label{fig:KnowChange_overview}
\end{figure*}

To instantiate this idea, we present a knowledge-guided change data synthesis framework for remote sensing, named KnowChange (Fig.~\ref{fig:KnowChange_overview}). Given pre-change images with rich semantic annotations and user-specified change types, KnowChange first prompts a pretrained VLM to infer plausible change locations and class transitions, producing a global layout of the post-change semantic mask. This change simulation eliminates the need for manually predefined transition rules, enabling a more diverse range of class transitions by reasoning over scene contexts. Moreover, desired change types are directly incorporated into the simulation process through prompts, avoiding repeated customization of transition rules across different application scenarios. 

The inferred layout is then instantiated by a layout-to-mask model, which refines the shapes of changed object and improves local object-context compatibility to produce pixel-level post-change semantic masks. Following existing synthesis pipelines, a mask-to-image model generates post-change images conditioned on the synthesized semantic masks. To support flexible synthesis of evolving change types, we curate a large-scale multi-category semantic segmentation dataset and develop effective training strategies for both models, allowing them to capture rich visual priors of object appearance. Consequently, KnowChange can generate valid object shapes and high-fidelity appearances for diverse change types without additional retraining.

\begin{figure}[t]
    \centering
    \includegraphics[width=0.7\linewidth]{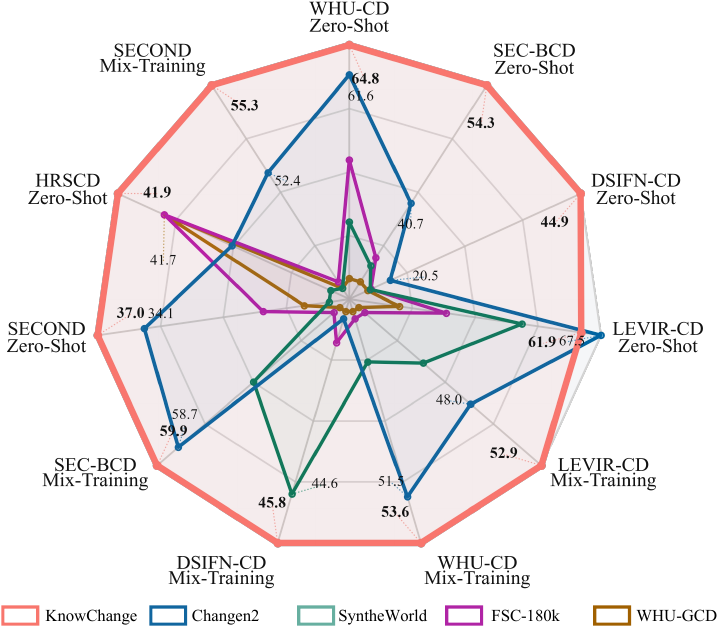}
    \caption{Downstream performance comparison of models trained with change data synthesized by different methods.}
    \label{fig:radar-plot}
\end{figure}

Using KnowChange, we synthesize three datasets, including Know-BCD, Know-SEC, and Know-HR for building and semantic change detection. Extensive experiments demonstrate that models trained on our synthetic datasets outperform those trained on existing ones (Fig.~\ref{fig:radar-plot}), achieving an average IoU gain of \textbf{6.64} on four building change detection benchmarks and an average F1 gain of \textbf{6.78} on two semantic change detection benchmarks. Furthermore, ablation studies show that knowledge-guided change simulation can be readily integrated into existing methods, such as HySCDG~\cite{benidir2025change} and Changen2~\cite{zheng2024changen2}, substantially improving the effectiveness of synthesized data for downstream change detection (Fig.~\ref{fig:cs_impact}).

Our contributions are summarized as follows:

\begin{itemize}

\item We introduce knowledge-guided change simulation that exploits pretrained VLMs to infer plausible change locations and class transitions, addressing the limited transition coverage of handcrafted rules and enhancing the diversity of synthesized change data.

\item We present KnowChange, a flexible change data synthesis framework that combines VLM-based change reasoning with generative models, enabling adaptive synthesis of user-specified change types without repeated customization of the synthesis pipeline.

\item We create three synthetic datasets for building and semantic change detection and demonstrate that training with our datasets improves detection accuracy and generalization over existing synthesis datasets.

\end{itemize}

\section{KnowChange}
\label{sec:mtds}

\subsection{Framework Overview}
Let $I_\text{pre} \in \mathbb{R}^{H \times W \times 3}$ and $S_\text{pre} \in \mathbb{R}^{H \times W}$ denote the pre-change image and its semantic mask, respectively, where each pixel in $S_\text{pre}$ indicates its semantic category. Given pre-change scene information ($I_\text{pre}$, $S_\text{pre}$) and a set of user-specified change types $\mathcal{C}=\{c_i\}_{i=1}^{N_c}$, KnowChange aims to synthesize the corresponding post-change image $I_\text{post} \in \mathbb{R}^{H \times W \times 3}$ and semantic mask $S_\text{post} \in \mathbb{R}^{H \times W}$. The binary change mask $M \in \{0,1\}^{H \times W}$ is obtained by comparing $S_\text{pre}$ and $S_\text{post}$. As illustrated in Fig.~\ref{fig:KnowChange_overview}, KnowChange consists of two key components: knowledge-guided change simulation and generalizable semantic-guided synthesis. The change simulation is performed by prompting a pretrained VLM to infer plausible change regions $r_i$ and corresponding post-change categories $c_i$, producing a global layout $\mathcal{G}_\text{chg}=\{(r_i,c_i)\}_{i=1}^{N_r}$. The semantic-guided synthesis component consists of a layout-to-mask (L2M) model and a mask-to-image(M2I) model. Given $S_\text{pre}$ and $\mathcal{G}_\text{chg}$, the L2M model generates a pixel-level post-change semantic mask $S_\text{post}$ by refining object shapes and ensuring local object-context compatibility. The M2I model then synthesizes the post-change image $I_\text{post}$ conditioned on $S_\text{post}$ and $I_\text{pre}$.

\subsection{Knowledge-Guided Change Simulation}
Existing rule-based simulation requires predefined class transitions, limiting the diversity of synthesized changes. We instead formulate change simulation as knowledge-guided reasoning, where a pretrained VLM infers plausible changes from scene contexts. Specifically, the VLM takes the pre-change image $I_\text{pre}$ and semantic mask $S_\text{pre}$ as inputs, along with a carefully designed prompt containing task description, desired change categories $\{c_i\}_{i=1}^{N_c}$, and the mapping between semantic classes and colors in $S_\text{pre}$ (detailed prompt design is provided in the Appendix). Guided by the prompt, the VLM conducts actual- and pseudo-change reasoning.

\textbf{Actual-Change Simulation:} Actual-change reasoning involves determining changed regions and their corresponding post-change categories ($\{(r_i,c_i)\}_{i=1}^{N_r}$). Depending on whether the post-change region preserves the shape of the pre-change region, we categorize actual changes into two complementary transition modes: shape-preserving and shape-altering transitions. In shape-preserving transitions, the shape of the original region is preserved, while its semantic category evolves from $c_\text{pre}$ to $c_i$. For example, structured farmland becoming abandoned may transition into grassland, where the original parcel boundary can be preserved. In this mode, the VLM predicts $c_i$, together with the corresponding pre-change category $c_\text{pre}$ and a region selection ratio $\alpha \in (0,1)$, from which $r_i$ is derived. Formally, let $\mathcal{K}=\{k_j\}_{j=1}^{N_k}$ denote the connected components of category $c_\text{pre}$ in $S_\text{pre}$, sorted in descending order of area. The number of selected components is determined as $n=\lceil\alpha N_k\rceil$, and the changed region is obtained by taking the union of the first $n$ components, \emph{i.e.}, $r_i=\bigcup_{j=1}^{n} k_j$. 

In contrast, shape-altering transitions require newly instantiated regions according to post-change categories. Typical examples include newly constructed buildings emerging on grassland, where no corresponding regions exist in the pre-change scene. To enable this transition, we provide the VLM with randomly generated candidate rectangular regions $\mathcal{B}=\{b_j\}_{j=1}^{N_b}$ in the prompt. The VLM then selects appropriate boxes and assigns post-change categories, producing $(r_i,c_i)$. Combining both transition modes yields the final change layout for post-change semantic mask synthesis.

\textbf{Pseudo-Change Simulation:} In real-world datasets, variations in imaging conditions may cause unchanged regions to exhibit visual differences across time while preserving their semantic categories  (\emph{e.g.}, grass appearing denser or sparser), a phenomenon known as pseudo-change. To improve synthesis realism, we simulate pseudo changes alongside actual changes. Specifically, the VLM identifies pre-change categories $c_\text{pre}$ that may exhibit pseudo changes and estimates their region perturbation ratios $\beta \in (0,1)$. Using the same region selection strategy as shape-preserving transitions, pseudo-change regions $r_\text{pse}$ are derived based on $\beta$. The obtained pairs $(r_\text{pse},c_\text{pre})$ are used to guide the M2I model to perform category-preserving appearance reconstruction during post-change image synthesis.

\subsection{Generalizable Semantic-Guided Synthesis}
The change simulation provides only a change layout specifying where and what changes occur, rather than the complete pixel-level post-change semantic mask required for image synthesis. The recent method~\cite{zan2025open} generates $S_\text{post}$ by sampling object shapes of post-change categories $c_i$ from manually maintained mask libraries and pasting them into changed regions $r_i$ of $S_\text{pre}$. However, such strategies are limited by shape diversity and cannot guarantee spatial coherence with surrounding areas, resulting in unrealistic scenes, \emph{e.g.}, newly constructed buildings may not seamlessly blend with adjacent land covers. We therefore introduce a semantic mask synthesis model to expand the sparse change layout into a complete post-change semantic mask, which subsequently guides image synthesis.

\textbf{Semantic Mask Synthesis:} The change layout is generated through two transition modes: shape-preserving and shape-altering transitions. Accordingly, the post-change semantic mask is obtained by integrating the masks derived from the two transition layouts. Let $\mathcal{M}_{\text{sp}}$ and $\mathcal{M}_{\text{sa}}$ denote the masks derived from the shape-preserving and shape-altering layouts, respectively. The shape-preserving mask $\mathcal{M}_{\text{sp}}$ is directly obtained by replacing the semantic labels of regions $r_i$ in $S_{\text{pre}}$ with their corresponding post-change categories $c_i$. In contrast, the shape-altering layout only provides coarse region constraints (\emph{i.e.}, candidate boxes). Therefore, we employ a layout-to-mask (L2M) model, which takes $S_{\text{pre}}$ with changed regions masked out as input, to instantiate these regions and generate $\mathcal{M}_{\text{sa}}$. Specifically, L2M model adopts the FLUX.1~\cite{labs2025flux} architecture, which is equipped with two text encoders. The T5 encoder~\cite{raffel2020exploring} provides rich representations for complex textual descriptions, while the CLIP text encoder~\cite{radford2021learning} provides category-level representations aligned with visual concepts. Accordingly, the category-color mapping in $S_{\text{pre}}$ is encoded by the T5 encoder, while the desired post-change categories are encoded by the CLIP text encoder. The resulting textual representations are used as conditioning signals to generate $\mathcal{M}_{\text{sa}}$. Finally, $S_{\text{post}}$ is obtained by replacing the corresponding regions in $\mathcal{M}_{\text{sa}}$ with $\mathcal{M}_{\text{sp}}$.

\textbf{Post-Change Image Synthesis:} The change mask $M$ is computed from the semantic difference between $S_{\text{pre}}$ and $S_{\text{post}}$, while the pseudo-change mask $M_\text{pse}$ is derived from the VLM-inferred pseudo-change regions $r_\text{pse}$. The two masks are combined to obtain the re-rendering mask $\mathcal{M}=M \bigcup M_\text{pse}$. The regions indicated by $\mathcal{M}$ are marked out from $I_{\text{pre}}$, which is then fed into a mask-to-image (M2I) diffusion model to synthesize $I_{\text{post}}$ conditioned on $S_{\text{post}}$. Since pseudo-change regions preserve semantic categories between $S_{\text{pre}}$ and $S_{\text{post}}$, the M2I model re-renders objects with unchanged categories while introducing subtle appearance variations. In contrast, actual change regions involve class transitions and require appearances consistent with the post-change categories. Existing change data synthesis methods~\cite{zang2025changediff} use RGB-encoded semantic masks as conditions for image synthesis. However, such designs require a fixed color assignment for each category, making them difficult to scale to scenarios with diverse categories, where assigning visually distinguishable colors to a large number of categories becomes impractical. To address this limitation, we employ a CLIP text encoder to generate spatially dense embeddings from semantic categories in $S_{\text{post}}$. These embeddings are further transformed by an adapter into control features for the diffusion model. The adapter consists of a $1\times1$ channel projection, three stride-2 convolutional blocks, and a zero-initialized output convolution.

\textbf{Training for Generalization:} Although semantic-guided synthesis enables flexible change data generation, its generalization ability is constrained by the diversity of training data. Existing methods~\cite{zheng2024changen2,benidir2025change,zang2025changediff} typically train synthesis models on datasets with limited category coverage, restricting their ability to synthesize varied change categories. To enhance category-level generalization, we curate a large-scale semantic segmentation corpus by consolidating existing datasets, including OpenEarthMap~\cite{xia2023openearthmap}, FLAIR~\cite{garioud2023flair}, Vaihingen~\cite{rottensteiner2012isprs}, Potsdam~\cite{rottensteiner2012isprs}, GID~\cite{tong2020land}, and SkySA~\cite{zhu2025skysense}, forming a corpus of 138K remote sensing images. Dataset statistics are provided in Table~\ref{tab:training_datasets}. These datasets provide over 1,000 object categories, enabling the L2M and M2I models to learn rich object priors. 

During training, masked inputs are prepared to match the inference inputs of the L2M and M2I models. Inspired by image editing practices~\cite{nichol2021glide,suvorov2022resolution}, we design dedicated masking procedures for the two models. For L2M training, masked semantic masks are generated using category-aware instance masking and random region masking. The text conditions describe either the masked category name or the dominant categories within the masked region, enabling the model to learn category-specific shape priors while maintaining spatial coherence across different categories. For M2I training, masked images are generated with three masking granularities: instance-level masking for category-specific appearance learning, region-level masking for cross-category boundary synthesis, and global-level masking for scene-level reconstruction. 

\begin{figure}[t]
    \centering
    \includegraphics[width=0.95\linewidth]{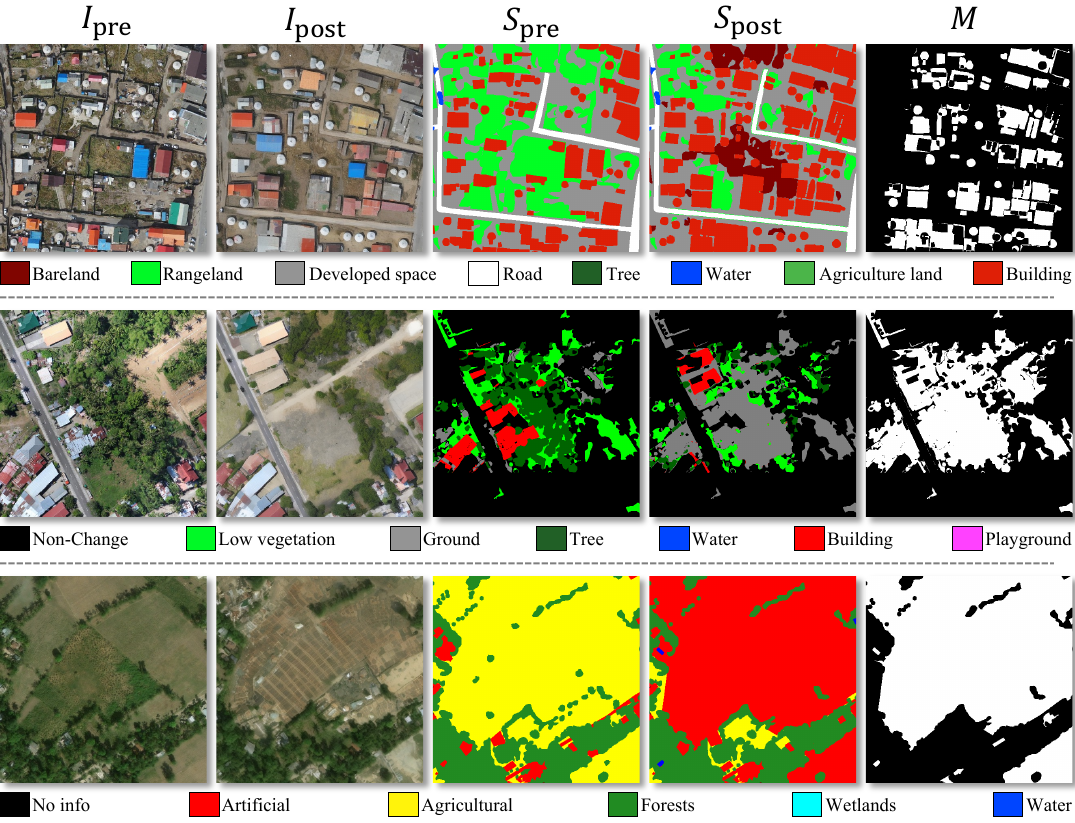}
    \caption{Examples from Know-BCD, Know-SEC, and Know-HR (top to bottom).}
    \label{fig:KnowChange_data_present}
\end{figure}

For $k\in\{S,I\}$, L2M and M2I are trained under a unified
conditional regression objective:
\begin{equation}
\mathcal{L}_{k}
=
\mathbb{E}_{(z_{0}^{k},\mathcal{C}_{k})\sim\mathcal{D}_{k},\,t,\,\epsilon}
\left[
\left\|
f_{\theta_{k}}(z_{t}^{k},t,\mathcal{C}_{k})-y_{k}
\right\|_{2}^{2}
\right],
\end{equation}
where $\mathcal{D}_{k}$ denotes the corresponding training
distribution. For L2M, $z_{0}^{s}$ denotes the latent representation
of the complete semantic mask $S_{\mathrm{pre}}$, and
$z_{t}^{s}=(1-t)z_{0}^{s}+t\epsilon$,
$f_{\theta_{s}}=v_{\theta_{s}}$,
$y_{s}=\epsilon-z_{0}^{s}$, and
$\mathcal{C}_{s}=({S}_{\mathrm{pre}},\mathcal{M},P)$,
yielding a flow-matching objective. For M2I,
$z_{0}^{I}$ is the latent of $I_{\mathrm{pre}}$,
$z_{t}^{I}$ is its noisy latent,
$f_{\theta_{I}}=\epsilon_{\theta_{I}}$,
$y_{I}=\epsilon$, and
$\mathcal{C}_{I}=
({z}_{0}^{I},\mathcal{M},
S_{\mathrm{post}})$,
yielding a noise-prediction objective. The M2I objective jointly
supervises ControlNet and the adapter to learn spatial and semantic
guidance, respectively. Here, $\mathcal{M}$ is sampled
using the three masking procedures described above, $t$ denotes the
normalized timestep, and
$\epsilon\sim\mathcal{N}(0,\mathbf{I})$.

\subsection{Flexible Change Data Synthesis}
Unlike existing methods that require manually designed transition rules and repeated retraining of synthesis models for different change types, a single trained KnowChange framework can flexibly synthesize change data with diverse change types. We combine OpenEarthMap and FLAIR as source datasets for change data synthesis. These datasets contain heterogeneous label taxonomies and provide 26 semantic categories in total. Based on this rich semantic annotation, KnowChange synthesizes three datasets for building and semantic change detection, namely Know-BCD, Know-SEC, and Know-HR. Know-SEC follows the change categories defined in SECOND~\cite{yang2021asymmetric}, while Know-HR follows those defined in HRSCD~\cite{daudt2019multitask}. Each dataset contains 10K samples, with examples shown in Fig.~\ref{fig:KnowChange_data_present}. Detailed statistics, comparisons with existing synthetic datasets, additional examples, and details of the synthesis process are provided in Appendix.

\section{Experiments}
\label{sec:exp}

\subsection{Experimental Setup}
\textbf{Datasets and Evaluation Metrics:} We evaluate the utility of synthesized data on six widely used change detection benchmarks, including four building change detection (BCD) datasets (LEVIR-CD~\cite{chen2020spatial}, WHU-CD~\cite{ji2018fully}, DSIFN-CD~\cite{zhang2020deeply}, and SEC-BCD~\cite{yang2021asymmetric}) and two semantic change detection (SCD) datasets (SECOND and HRSCD). We report F1-score and IoU for BCD, and F1-score, mIoU, SCS~\cite{toker2022dynamicearthnet} and SeK~\cite{yang2021asymmetric} for SCD.

\textbf{Downstream Models:} For downstream evaluation, we train representative change detection models on synthesized datasets and test them on real-world benchmarks. Specifically, we adopt ChangeFormer~\cite{bandara2022transformer} for BCD and Change3D~\cite{zhu2025change3d} for SCD, trained for 42K/30K iterations with batch sizes of 24/8, respectively.

\textbf{Synthesis Model Details:} The L2M adopts FLUX.1-Fill as the backbone and is fine-tuned with LoRA~\cite{hu2022lora} (rank 32) using AdamW for 30 epochs with a batch size of 16 and a learning rate of $5 \times 10^{-5}$. For the M2I model, we adopt an SD-v1.5~\cite{rombach2022high} model fine-tuned on remote sensing images~\cite{benidir2025change}, and replace the image-conditioned encoder with a CLIP text encoder. The U-Net and ControlNet~\cite{zhang2023adding} are optimized with learning rates of $2 \times 10^{-5}$ and $5 \times 10^{-6}$, respectively. The M2I model is trained for 20 epochs with a batch size of 32. Both models are trained on the collected 138K samples with $512\times512$ inputs using two NVIDIA 96G H20 GPUs. More training details are provided in Appendix. 

\begin{table}[t]
\centering
\caption{Summary of remote sensing datasets used for training the layout-to-mask and mask-to-image models. }
\label{tab:training_datasets}
\resizebox{0.8\linewidth}{!}{%
\begin{tabular}{l|c|c|c|c}
\toprule
Dataset & Resolution (m) & Bands & \# Classes & \# Samples \\
\midrule
OpenEarthMap\cite{xia2023openearthmap} & 0.25 $\sim$ 0.5 & RGB & 8 & 11,662 \\
FLAIR\cite{garioud2023flair} & 0.2 & RGB + NIR & 19 & 61,712 \\
Vaihingen\cite{rottensteiner2012isprs} & 0.09 & IR-R-G & 6 & 864 \\
Potsdam\cite{rottensteiner2012isprs} & 0.05 & RGB + IR & 6 & 7,406 \\
GID\cite{tong2020land} & 1 & RGB + NIR & 15 & 23,100 \\
SkySA\cite{zhu2025skysense} & Variable & RGB & 1,763 & 33,775 \\
\bottomrule
\end{tabular}
}%
\end{table}

\subsection{Downstream Utility of Synthesized Data}

\begin{table*}[ht]
\centering
\caption{Synthetic-to-real transfer results on building change detection benchmarks. \textit{Average} indicates the mean performance over four benchmarks. The best and second-best results are shown in bold and underline, respectively.}
\label{tab:comparison}
\setlength{\tabcolsep}{1.8mm}
\resizebox{\textwidth}{!}{%
\begin{tabular}{ccccccccccccc}
\toprule
\multirow{2}{*}{Dataset} & \multirow{2}{*}{Source Task} & \multirow{2}{*}{\# Samples} & \multicolumn{2}{c}{LEVIR-CD} & \multicolumn{2}{c}{WHU-CD} & \multicolumn{2}{c}{DSIFN-CD} & \multicolumn{2}{c}{SEC-BCD} & \multicolumn{2}{c}{Average} \\  \cmidrule(lr){4-5}\cmidrule(lr){6-7}\cmidrule(lr){8-9}\cmidrule(lr){10-11}\cmidrule(lr){12-13}
 & & & IoU & F1 & IoU & F1 & IoU & F1 & IoU & F1 & IoU & F1 \\ \midrule
WHU-GCD & SCD & 25K & 1.26 & 2.48 & 1.67 & 3.30 & 2.91 & 5.66 & 5.36 & 10.19 & 2.80 & 5.40 \\

Changen2-S9 & SCD & 27K & 1.99 & 3.90 & 3.80 & 7.32 & 4.73 & 9.04 & 6.69 & 12.54 & 4.30 & 8.20 \\

FSC-180k  & SCD & 180K & 11.19 & 20.13 & 33.50 & 50.19 & 5.49 & 10.40 & 16.40 & 28.18 & 16.64 & 27.22 \\

\textbf{Know-SEC} & SCD & \textbf{10K} & 30.81 & 47.10 & 34.95 & 51.79 & \textbf{32.38} & \textbf{48.92} & \underline{29.54} & \underline{45.61} & 31.92 & \underline{48.36} \\ \midrule

SyntheWorld & BCD & 40K & 28.66 & 44.55 & 23.11 & 37.54 & 5.08 & 9.66 & 14.23 & 24.91 & 17.77 & 29.17 \\
Changen2-S1 & BCD & 15K & \textbf{50.89} & \textbf{67.45} & \underline{44.48} & \underline{61.57} & 11.40 & 20.48 & 25.52 & 40.67 & \underline{33.07} & 47.54 \\
\textbf{Know-BCD} & BCD & \textbf{10K} & \underline{44.81} & \underline{61.89} & \textbf{47.87} & \textbf{64.75} & \underline{28.93} & \underline{44.88} & \textbf{37.24} & \textbf{54.27} & \textbf{39.71} & \textbf{56.44} \\
\bottomrule
\end{tabular}%
}
\end{table*}

\begin{table*}[ht]
\centering
\caption{Synthetic-to-real transfer results on semantic change detection benchmarks. \textit{Average} indicates the mean performance over two benchmarks. The best and second-best results are shown in bold and underline, respectively.}
\label{tab:second_hrscd}
\setlength{\tabcolsep}{1.8mm}
\resizebox{\textwidth}{!}{%
\begin{tabular}{cccccccccccc}
\toprule
\multirow{2}{*}{Dataset} & \multirow{2}{*}{\#Samples} & \multicolumn{3}{c}{SECOND} & \multicolumn{3}{c}{HRSCD} & \multicolumn{3}{c}{Average} & \multirow{2}{*}{$\Delta$ F1 (vs. ours)} \\
\cmidrule(lr){3-5}\cmidrule(lr){6-8}\cmidrule(lr){9-11}
& & F1 & mIoU & SCS & F1 & mIoU & SCS & F1 & mIoU & SCS & \\
\midrule
WHU-GCD & 25K & 15.32 & 19.85 & 18.06 & \underline{41.71} & 39.10 & \textbf{36.12} & 28.52 & 29.48 & 27.09 & -10.97 \\
FSC-180k & 180K & 22.95 & 50.40 & 23.84 & 39.36 & \textbf{53.98} & 31.46 & 31.16 & \underline{52.19} & \underline{27.50} & -8.33 \\
Changen2-S9 & 27K & \underline{34.06} & \textbf{57.75} & \underline{26.70} & 31.35 & 41.59 & 17.81 & \underline{32.71} & 49.67 & 22.26 & -6.78 \\
\textbf{Know-SEC/HR} & \textbf{10K} & \textbf{37.03} & \underline{55.15} & \textbf{28.31} & \textbf{41.94} & \underline{50.78} & \underline{32.16} & \textbf{39.49} & \textbf{52.97} & \textbf{30.24} & - \\
\bottomrule
\end{tabular}%
}
\end{table*}

\begin{table*}[ht]
\centering
\caption{Performance comparison of synthetic data augmentation with 5\% real training data from different benchmarks. For BCD tasks, models are evaluated on four BCD benchmarks, and results are reported as average IoU and F1. For SECOND, models are evaluated on the SECOND test set. The best and second-best results are shown in bold and underline, respectively.}
\label{tab:mix_training_bcd_second}
\setlength{\tabcolsep}{1.15mm}
\resizebox{\textwidth}{!}{%
\begin{tabular}{ccccccccc|ccccc}
\toprule
\multirow{2}{*}{Dataset} & \multicolumn{2}{c}{5\% LEVIR-CD} & \multicolumn{2}{c}{5\% WHU-CD} & \multicolumn{2}{c}{5\% DSIFN-CD} & \multicolumn{2}{c|}{5\% SEC-BCD} & \multirow{2}{*}{Dataset} & \multicolumn{4}{c}{5\% SECOND} \\
\cmidrule(lr){2-3}\cmidrule(lr){4-5}\cmidrule(lr){6-7}\cmidrule(lr){8-9}\cmidrule(lr){11-14}
& IoU & F1 & IoU & F1 & IoU & F1 & IoU & F1 & & F1 & mIoU & SeK & SCS \\
\midrule
FSC-180k & 15.80 & 26.32 & 17.81 & 29.77 & 21.96 & 35.84 & 21.96 & 33.56 & No Syn. & 50.84 & 65.05 & 10.53 & 38.81  \\
SyntheWorld & 28.97 & 43.27 & 29.02 & 41.75 & \underline{28.95} & \underline{44.62} & 36.09 & 52.79 & FSC-180k & 41.34 & 60.00 & 4.79 & 26.95 \\
Changen2-S1 & \underline{32.61} & \underline{48.01} & \underline{36.62} & \underline{51.54} & 17.49 & 29.74 & \underline{42.56} & \underline{58.66} & Changen2-S9 & \underline{52.37} & \underline{66.21} & \underline{11.85} & \textbf{39.93} \\
\textbf{Know-BCD} & \textbf{36.28} & \textbf{52.88} & \textbf{37.13} & \textbf{53.59} & \textbf{29.74} & \textbf{45.77} & \textbf{43.20} & \textbf{59.94} & \textbf{Know-SEC} & \textbf{55.27} & \textbf{67.24} & \textbf{13.89} & \underline{39.87} \\
\bottomrule
\end{tabular}%
}
\end{table*}

\textbf{Synthetic-to-Real Transfer:} We train change detection models solely on synthesized datasets and evaluate their generalization on real-world benchmarks. Table~\ref{tab:comparison} reports synthetic-to-real transfer results on BCD benchmarks. Following prior practices~\cite{song2024syntheworld}, we filter building-related samples from synthesized datasets originally designed for SCD and use them for BCD model training. Among the four SCD-oriented synthesized datasets, the Know-SEC-trained model generalizes best to real-world BCD benchmarks despite using the fewest training samples, outperforming models trained with other SCD-oriented datasets by over 15/21 points in average IoU/F1. Among the three BCD-oriented synthesized datasets, training with Know-BCD achieves the best transfer results with only 10K samples. Although lower than Changen2-S1~\cite{zheng2024changen2} on LEVIR-CD, it substantially outperforms Changen2-S1 on the other three benchmarks, yielding an average IoU gain of 6.64 points. We further extend the evaluation to SCD benchmarks. Table~\ref{tab:second_hrscd} reports the results. Existing methods rely on handcrafted transition rules, limiting change diversity and introducing benchmark-specific biases. Consequently, models trained on synthetic data may generalize well to one benchmark but fail on another, as observed on WHU-GCD~\cite{zan2025open}. In contrast, KnowChange enables flexible benchmark-specific synthesis, yielding customized training data for SECOND and HRSCD. Models trained on our synthesized datasets outperform those trained on existing ones, achieving an average F1 improvement of 6.78 points.

\textbf{Synthetic Data Augmentation:} Beyond direct synthetic-to-real transfer, we evaluate synthesized data augmentation under limited real-data regimes. Each synthetic dataset is combined with 5\% real training data from one BCD benchmark at a time, and the resulting models are evaluated on all four BCD benchmarks. As shown in Table~\ref{tab:mix_training_bcd_second}, models trained with Know-BCD consistently outperform those trained with other synthetic datasets across all four real-data settings. The largest gains reach 3.67 and 4.87 points in average IoU and F1, respectively. We further conduct this augmentation experiment on the SECOND benchmark. As shown in Table~\ref{tab:mix_training_bcd_second}, Know-SEC achieves the best augmentation performance among existing synthetic datasets. Overall, the improvements in both transfer and augmentation experiments validate the effectiveness of KnowChange for generating high-quality synthetic change data.

\begin{figure}[ht]
    \centering
    \includegraphics[width=0.95\linewidth]{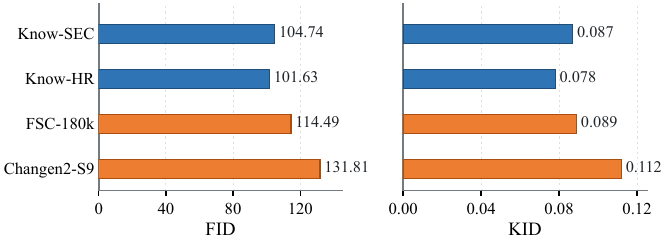}
    \caption{FID and KID comparison of synthetic datasets.}
    \label{fig:FID-KID}
\end{figure}

\textbf{Synthetic Data Quality Analysis:} To understand the performance gains, we assess synthesized data quality using FID and KID metrics, with SECOND as the reference distribution. As shown in Fig.~\ref{fig:FID-KID}, KnowChange-generated data achieves lower FID and KID scores than existing synthetic data, indicating better alignment with real-world change data.

\subsection{Ablation Studies}
\textbf{Effect of Knowledge-Guided Change Simulation:} Since generative models are essential components of the synthesis pipeline but are not the focus of our contribution, we keep them fixed to isolate the effect of knowledge-guided change simulation (Table~\ref{tab:ablation_component}). For building change synthesis, we first establish a baseline using random copy-paste to generate post-change semantic masks. This strategy often produces unrealistic changes, \emph{e.g.}, pasting buildings onto water, leading to an average IoU of only 3.96. We then replace copy-paste with shape-altering transition (\textit{SAT}). Although \textit{SAT} cannot fully eliminate change location errors, the L2M model improves object-context compatibility, \emph{e.g.}, generating buildings near water boundaries rather than directly on water surfaces, increasing average IoU to 37.58. Adding shape-preserving transition (\textit{SPT}) provides marginal gains, mainly because its generated building-to-building transitions are rarely considered in existing BCD benchmarks. Incorporating pseudo-change simulation (\textit{PCS}) further improves average IoU by over 6.5 points. Finally, pretrained VLMs for knowledge-guided reasoning further boost the average IoU to 46.34. On SCD, removing VLM guidance causes a 9.34-point F1 drop, validating the importance of knowledge-guided reasoning for realistic and diverse change simulation.

\begin{table}[ht]
    \centering
        \caption{Ablation study of knowledge-guided change simulation. Generative models are fixed, and different components are progressively added to evaluate their contributions.}
    \label{tab:ablation_component}
\setlength{\tabcolsep}{2mm}
    \begin{tabular}{cccc}
    \toprule
    \multirow{2}{*}{Method} & LEVIR-CD& WHU-CD & \multirow{2}{*}{Average} \\
    & (IoU) & (IoU) & \\ \midrule
    Copy-Paste & 4.10 & 3.82 & 3.96 \\ \midrule
    \textit{SAT} & 33.95 & 41.20 & 37.58 \\
    + \textit{SPT} & 35.33 & 40.31 & 37.82 \\
    + \textit{PCS} & 44.07 & 44.62 & 44.35 \\
    + \textit{VLM} & \textbf{44.81} & \textbf{47.87} & \textbf{46.34} \\ \midrule
     \multicolumn{4}{c}{SECOND (F1/mIoU/SCS)} \\ \midrule
     \textit{SAT}+\textit{SPT}+\textit{PCS} &  24.07 & 49.93 & 22.97 \\
     +\textit{VLM} & \textbf{33.41} & \textbf{50.39} & \textbf{29.50} \\
     \bottomrule
    \end{tabular}
\end{table}

\textbf{Impact of VLMs:} To investigate the impact of different VLMs on change simulation, we replace the reasoning model with different pretrained VLMs and evaluate the resulting synthesized data on the SECOND benchmark. As shown in Table~\ref{tab:VLM-impact}, all VLM-guided strategies consistently outperform the w/o VLM baseline. Among different VLMs, Qwen3-VL~\cite{Qwen3-VL} achieves the best F1 (33.41) and SCS (29.50), while DouBao-2.0-mini~\cite{seed2026seed2} and GLM-4.6V~\cite{hong2025glm} obtain slightly higher mIoU. These results show that pretrained VLMs serve as effective knowledge providers for change simulation, and VLM selection influences the quality of synthesized data. We adopt Qwen3-VL as the default reasoning model in all experiments.

\begin{table}[ht]
    \centering
        \caption{Impact of VLMs on knowledge-guided change simulation evaluated on the SECOND benchmark.}
    \label{tab:VLM-impact}
\setlength{\tabcolsep}{0.9mm}
    \begin{tabular}{ccccc}
    \toprule
    VLM Guidance & F1 & mIoU & SCS & $\Delta$F1 (w/o VLM) \\ \midrule
    w/o VLM & 24.07 & 49.93 & 22.97 & - \\
    GLM-4.6V & 29.84 & \underline{53.16} & 25.90 & 5.77 \\
    DouBao-2.0-mini & \underline{31.72} & \textbf{53.66} & \underline{25.98} & \underline{7.65} \\
    Qwen3-VL & \textbf{33.41} & 50.39 & \textbf{29.50} & \textbf{9.34} \\
    \bottomrule
    \end{tabular}
\end{table}

\begin{figure}
    \centering
    \includegraphics[width=0.95\linewidth]{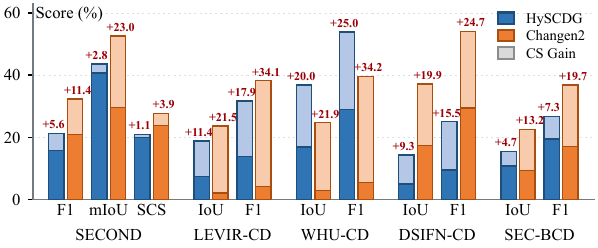}
    \caption{Performance comparison of existing synthesis methods with and without our change simulation.}
    \label{fig:cs_impact}
\end{figure}

\textbf{Plug-and-Play Change Simulation:} We integrate our knowledge-guided change simulation with existing synthesis methods, including HySCDG and Changen2, by replacing their change mask generation while keeping image generation models unchanged. As shown in Fig.~\ref{fig:cs_impact}, this replacement improves the performance of models trained on synthesized data across both BCD and SCD tasks. The improvement is particularly significant when combined with Changen2, achieving a 23-point mIoU gain on SECOND and over 21-point IoU gains on both LEVIR-CD and WHU-CD. These results demonstrate the potential of our change simulation as a plug-and-play component for existing synthesis methods.

\begin{figure}[t]
    \centering
    \begin{subfigure}[t]{0.32\linewidth}
        \centering
        \includegraphics[width=\linewidth]{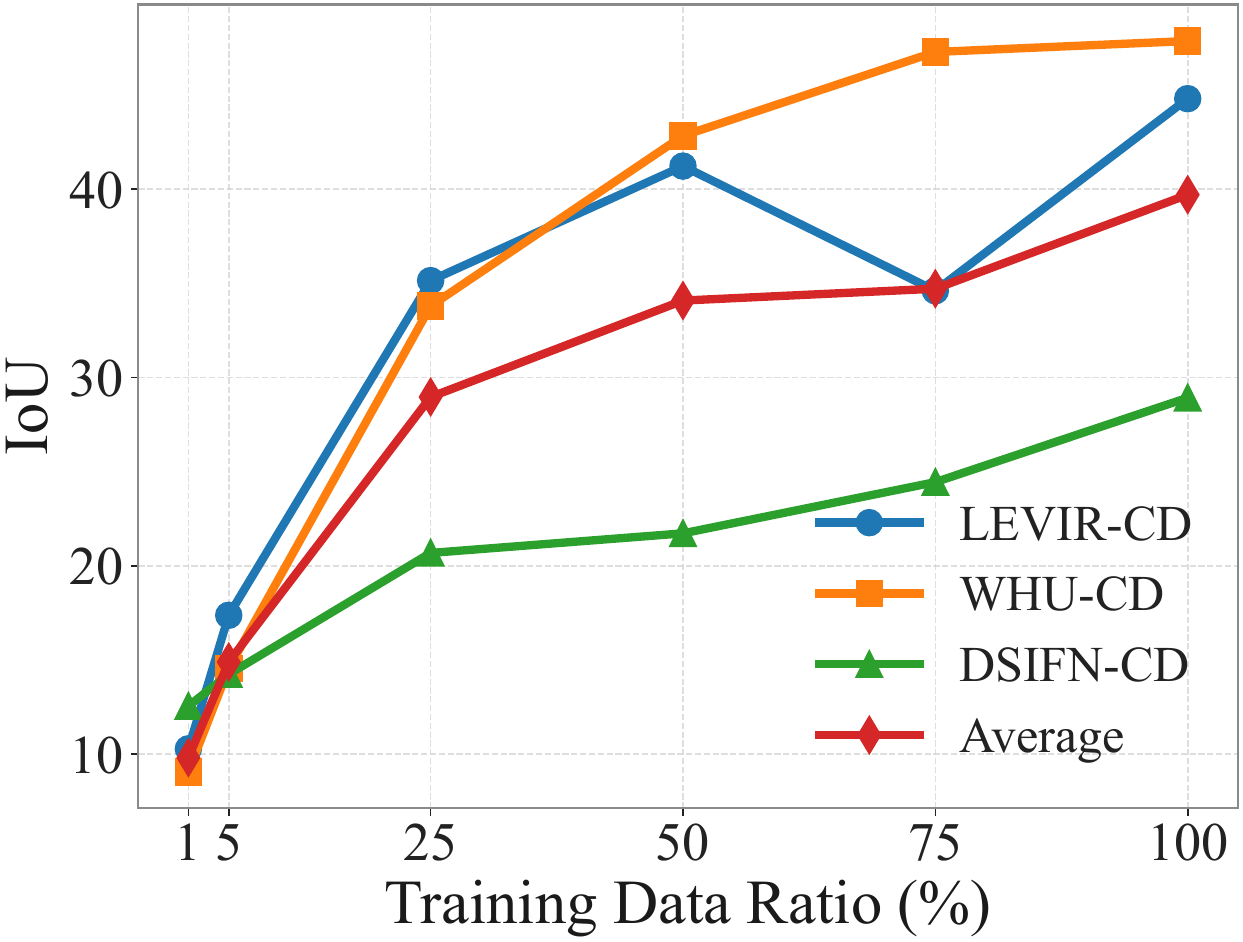}
    \end{subfigure}\hfill
    \begin{subfigure}[t]{0.32\linewidth}
        \centering
        \includegraphics[width=\linewidth]{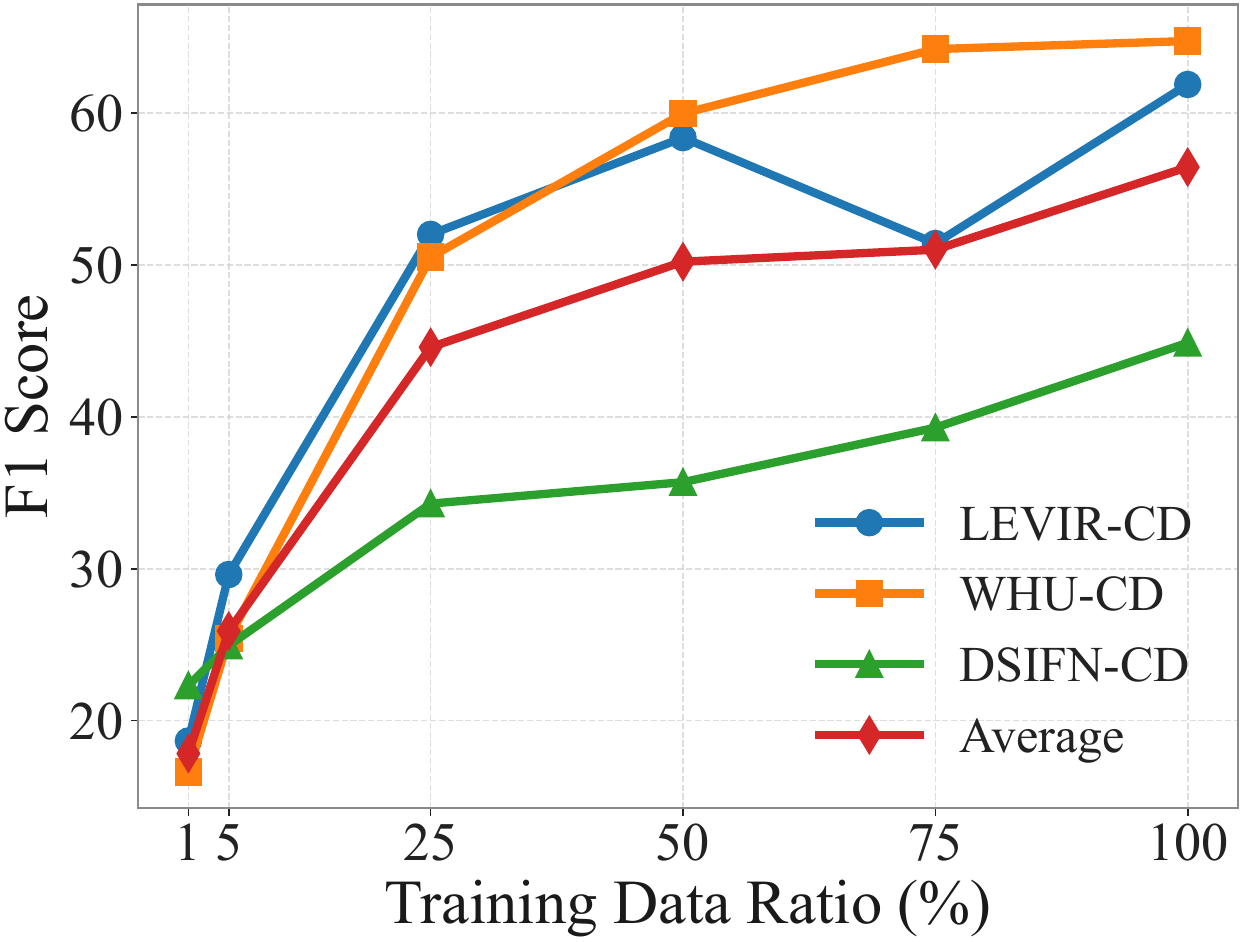}
    \end{subfigure}\hfill
    \begin{subfigure}[t]{0.32\linewidth}
        \centering
        \includegraphics[width=\linewidth]{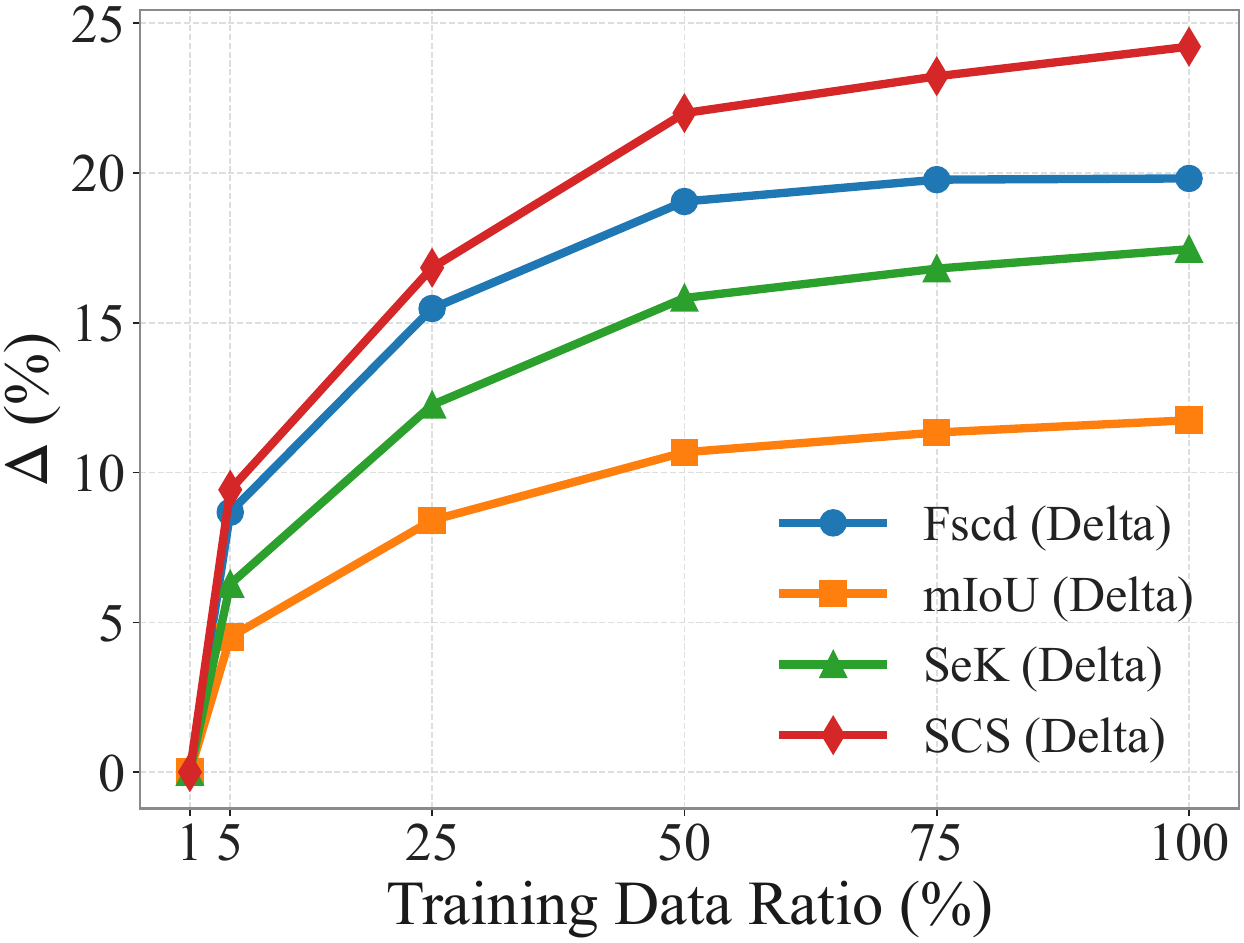}
    \end{subfigure}
    \caption{Scaling analysis of synthetic training data. Left and middle: Know-BCD scaling results on binary change detection benchmarks. Right: relative improvements on SECOND under varying Know-SEC training data ratios.}
    \label{fig:abl_data_quantity}
\end{figure}

\textbf{Scaling Analysis of Synthetic Data:} We train models with different amounts of Know-BCD, ranging from 1\% to 100\%, and evaluate them on four BCD benchmarks. As shown in Fig.~\ref{fig:abl_data_quantity}, increasing synthesized data from 1\% to 5\% brings the most significant performance gains across four BCD benchmarks, suggesting that even a small amount of synthesized data provides valuable information for learning change patterns. Further increasing synthesized data consistently improves performance on three benchmarks, demonstrating the effectiveness of synthetic data scaling. On LEVIR-CD, although a performance drop is observed with 75\% data, the performance recovers with the full dataset. Overall, these results demonstrate that Know-BCD provides increasingly effective training signals as more synthesized data becomes available. The SECOND scaling results show similar trends: increasing Know-SEC consistently improves semantic change detection metrics, indicating that the scaling benefit extends from binary to semantic change detection.

\FloatBarrier

\section{Conclusion}

In this work, we introduce KnowChange, a knowledge-guided change data synthesis framework for remote sensing. Instead of using rule-based change simulation, KnowChange leverages pretrained vision-language models as knowledge sources to reason about plausible change regions and class transitions from scene contexts and desired changes. KnowChange combines knowledge-guided change simulation with generalizable layout-to-mask and mask-to-image models, enabling flexible synthesis of diverse change types. This process results in three synthetic datasets for building and semantic change detection. Extensive experiments demonstrate that KnowChange-generated data consistently improves synthetic-to-real transfer and synthetic data augmentation over existing synthetic datasets. Ablation studies highlight the importance of knowledge-guided reasoning and show that the proposed simulation can be seamlessly integrated into existing synthesis methods in a plug-and-play manner.

\clearpage
{\small
\bibliographystyle{unsrt}
\bibliography{ref}
}

\clearpage
\appendix
\clearpage
\hypersetup{pageanchor=false}
\setcounter{page}{1}
\appendix

\section*{Contents}
\noindent
A. \hspace{0.2cm} Change Data Synthesis Details \dotfill \pageref{synthtic-data-details} \\
B. \hspace{0.2cm} Additional Experimental Details \dotfill \pageref{experimental-details} \\
C. \hspace{0.2cm} Additional Experimental Results and Analysis \dotfill \pageref{additional-results} \\
D. \hspace{0.2cm} Related Work \dotfill \pageref{related-work} \\

\section{Change Data Synthesis Details}
\label{synthtic-data-details}

\subsection{Prompt Design for Global Layout Reasoning}
The knowledge-guided change simulation module leverages Qwen3-VL~\cite{Qwen3-VL} as a real-world knowledge source to infer plausible changes from pre-change scene contexts and user-specified change types. Given the pre-change image $I_\text{pre}$, semantic mask $S_\text{pre}$, and desired change categories $\mathcal{C}=\{c_i\}_{i=1}^{N_c}$, the VLM reasons about changed regions and their corresponding post-change categories, producing a change layout $\{(r_i,c_i)\}_{i=1}^{N_r}$. To support different change behaviors in remote sensing scenes, we design structured prompts for two actual-change transition modes, namely shape-preserving and shape-altering transitions, as well as pseudo-change simulation. All prompts use a unified JSON output format to facilitate reliable parsing and subsequent synthesis.

(1)~For shape-preserving transitions, the shape of the original region is preserved, while its semantic category evolves from $c_\text{pre}$ to $c_i$. The prompt takes as input $I_\text{pre}$, $S_\text{pre}$, the semantic category-color mapping, and desired change categories $\mathcal{C}$. The VLM predicts the post-change category $c_i$, together with the corresponding pre-change category $c_\text{pre}$ and a region selection ratio $\alpha \in (0,1)$, from which the changed region $r_i$ is derived. Formally, let $\mathcal{K}=\{k_j\}_{j=1}^{N_k}$ denote the connected components of category $c_\text{pre}$ in $S_\text{pre}$, sorted in descending order of area. The number of selected components is determined as $n=\lceil\alpha N_k\rceil$, and the changed region is obtained by taking the union of the first $n$ components, \emph{i.e.}, $r_i=\bigcup_{j=1}^{n} k_j$. An example of the prompt and output for shape-preserving transitions is shown below.
\begin{center}
\includegraphics[width=0.72\linewidth,height=0.30\textheight,keepaspectratio]{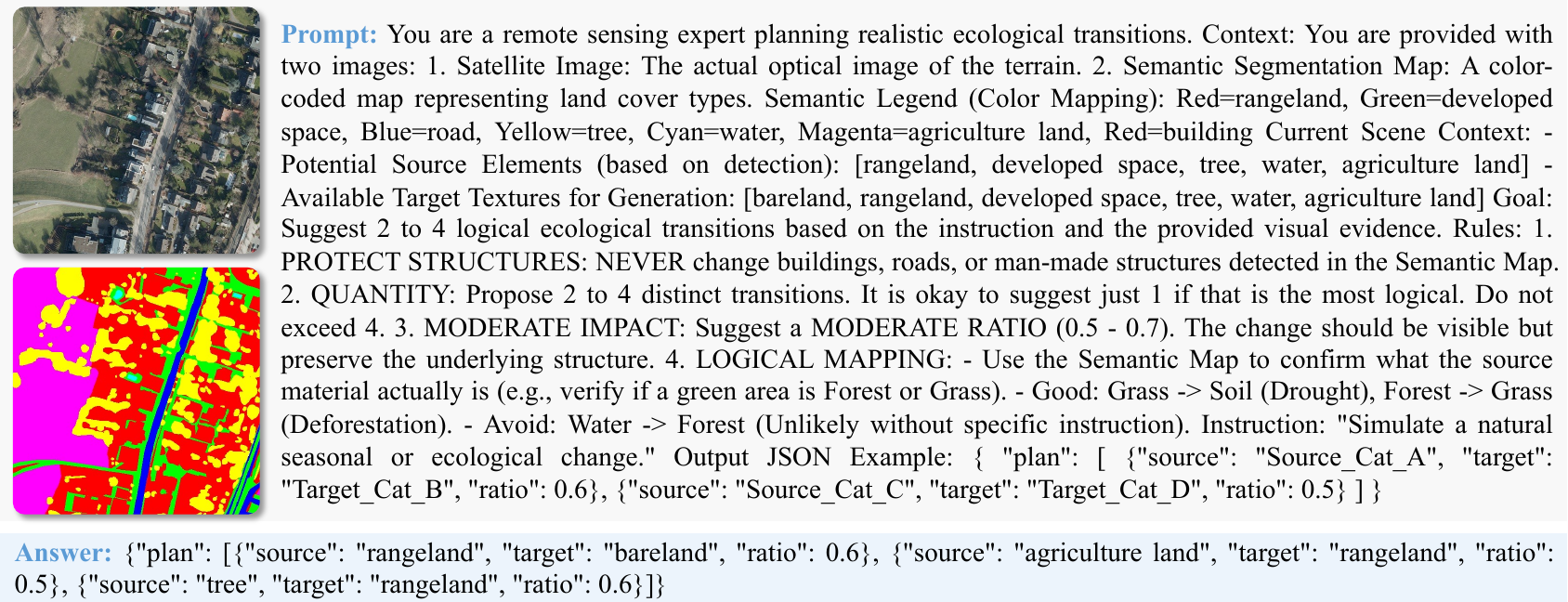}
\captionof{figure}{Example prompt and VLM output for shape-preserving transitions.}
\label{fig:shape_preserve_prompt}
\end{center}

(2)~For shape-altering transitions, newly instantiated regions are generated according to post-change categories, and their shapes may differ from existing pre-change regions. The prompt takes as input $I_\text{pre}$, $S_\text{pre}$, the semantic category-color mapping, randomly generated candidate rectangular regions $\mathcal{B}=\{b_j\}_{j=1}^{N_b}$, and desired change categories $\mathcal{C}$. The VLM selects appropriate candidate boxes and assigns post-change categories, producing $(r_i,c_i)$. These coarse regions are subsequently refined by the layout-to-mask model into pixel-level object shapes that are compatible with the surrounding scene context. An example of the prompt and output for shape-altering transitions is shown below.

\begin{center}
\includegraphics[width=0.72\linewidth,height=0.30\textheight,keepaspectratio]{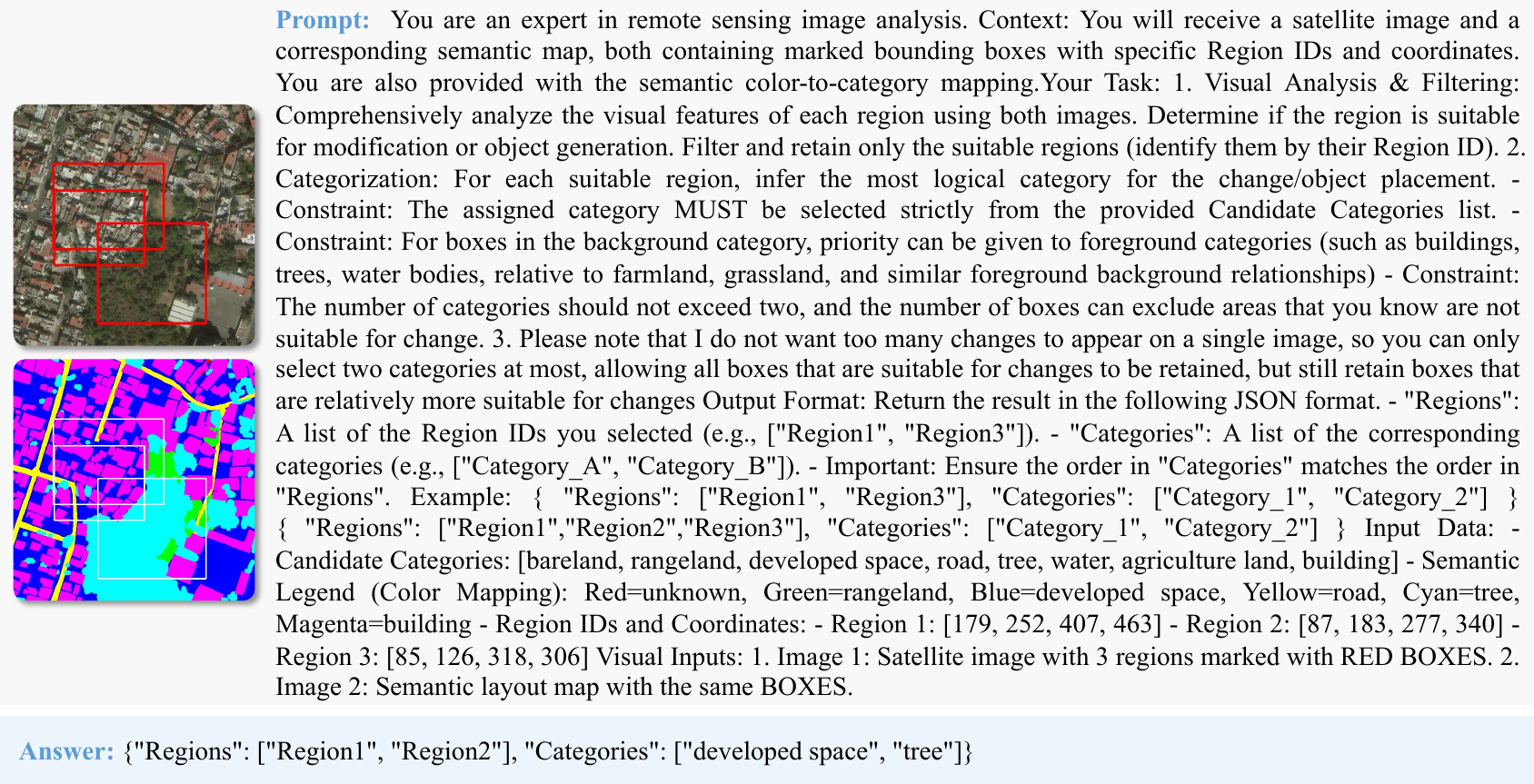}
\captionof{figure}{Example prompt and VLM output for shape-altering change simulation.}
\label{fig:shape_alter_prompt}
\end{center}

(3)~For pseudo-change simulation, the semantic category remains unchanged while the visual appearance varies due to imaging-condition differences such as illumination, season, atmosphere, or sensor variation. The prompt takes as input $I_\text{pre}$, $S_\text{pre}$, the semantic category-color mapping, and the categories present in the image. The VLM identifies pre-change categories $c_\text{pre}$ that may exhibit pseudo changes and estimates their region perturbation ratios $\beta \in (0,1)$. For each predicted category $c_\text{pre}$, let $\mathcal{K}=\{k_j\}_{j=1}^{N_k}$ denote its connected components in $S_\text{pre}$, sorted in descending order of area. The number of selected pseudo-change components is determined as $n=\lceil\beta N_k\rceil$, and the pseudo-change region is obtained by taking the union of the first $n$ components, \emph{i.e.}, $r_\text{pse}=\bigcup_{j=1}^{n} k_j$. The obtained pairs $(r_\text{pse},c_\text{pre})$ are used to guide the M2I model to perform category-preserving appearance reconstruction during post-change image synthesis. An example of the prompt and output for pseudo-change simulation is shown below.

\begin{center}
\includegraphics[width=0.72\linewidth,height=0.30\textheight,keepaspectratio]{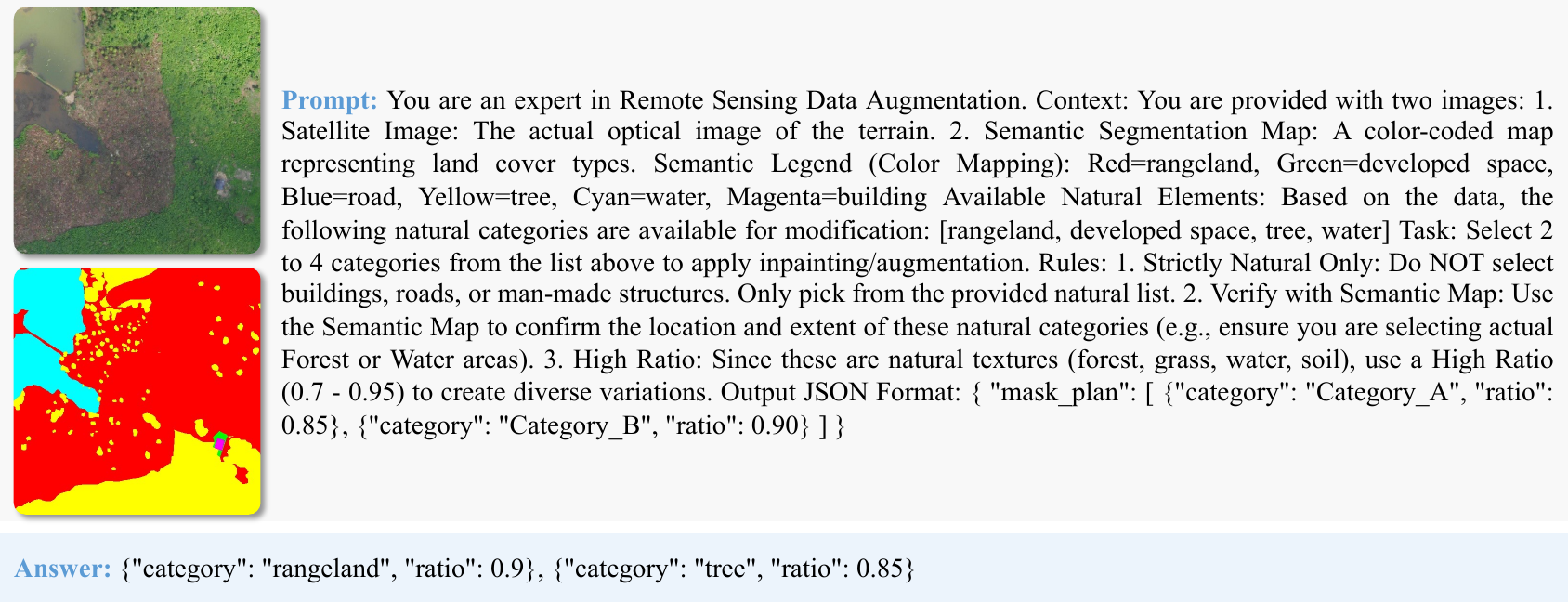}
\captionof{figure}{Example prompt and output for pseudo-change simulation.}
\label{fig:pseudo_change_prompt}
\end{center}

\subsection{More Examples}
\begin{figure}[!t]
    \centering
    \includegraphics[width=0.84\linewidth]{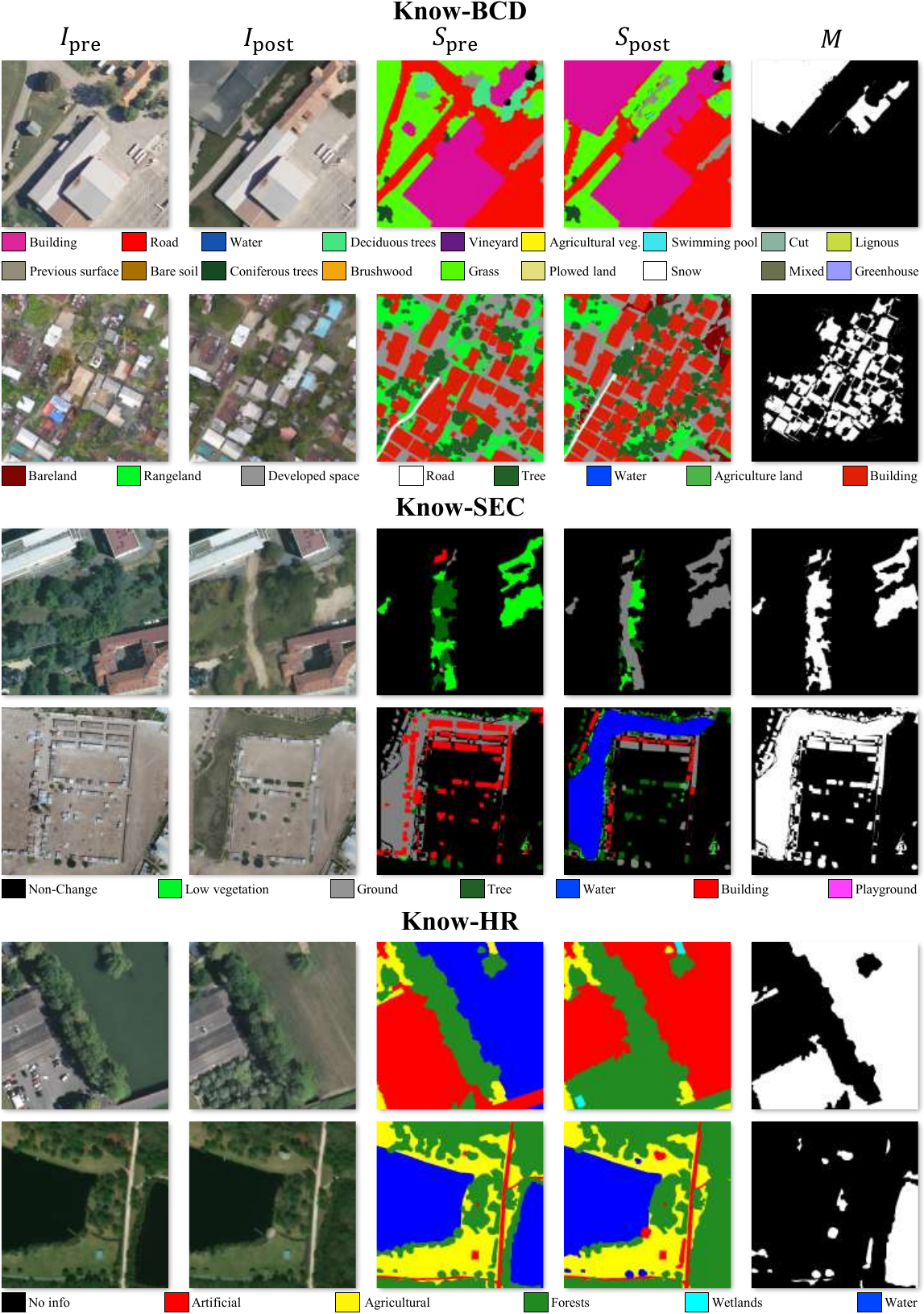}
    \caption{Visualization of generated samples from our three synthetic datasets: Know-BCD, Know-SEC, and Know-HR. In each dataset, the first row uses source images from FLAIR and the second row uses source images from OpenEarthMap.}
    \label{fig:KnowChange_example}
\end{figure}

 We present additional examples from our three generated datasets: Know-BCD (building change detection), Know-SEC (semantic change detection based on SECOND categories), and Know-HR (semantic change detection based on HRSCD categories). All datasets are synthesized from two source benchmarks: OpenEarthMap~\cite{xia2023openearthmap} and FLAIR~\cite{garioud2023flair}. The examples organized by source dataset are shown in Fig.~\ref{fig:KnowChange_example}.

\section{Additional Experimental Details}
\label{experimental-details}

\subsection{Data Source}

\begin{table*}[htbp]
    \centering
        \caption{Detailed statistics of the datasets used for binary change detection (BCD) and semantic change detection (SCD). Train/Val/Test denotes the number of samples after preprocessing.}
    \label{tab:datasets_detail}
\setlength{\tabcolsep}{4mm}
    \resizebox{\textwidth}{!}{%
\begin{tabular}{l c c c c c}
        \toprule
        Dataset & Resolution (m) & \# Samples & Image Size
        & Train/Val/Test & Task \\
        \midrule
        LEVIR-CD & 0.5 & 637
        & $1024 \times 1024$ & 2,548/--/1,392 & BCD \\
        WHU-CD & 0.075 & 1
        & $32507 \times 15354$ & 1,260/--/690 & BCD \\
        DSIFN-CD & 0.03--1 & 3,940
        & $512 \times 512$ & 3,600/340/48 & BCD \\
        SECOND & 0.5--3 & 4,662
        & $512 \times 512$ & 2,968/--/1,694 & BCD \& SCD \\
        HRSCD & 0.5 & 291
        & $10000 \times 10000$ & 905/116/123 & SCD \\
        \bottomrule
    \end{tabular}%
}
\end{table*}

\textbf{Evaluation Datasets for Change Detection:}
To comprehensively assess the utility of synthetic data, we evaluate on five widely adopted change detection benchmarks, covering both Binary Change Detection (BCD) and Semantic Change Detection (SCD) tasks. Detailed statistics of the original datasets are provided in Table~\ref{tab:datasets_detail}. These datasets collectively span diverse geographic regions, seasonal variations, and change patterns, enabling rigorous evaluation of model generalization. To ensure consistent evaluation across heterogeneous benchmarks, we standardize all input images to $512 \times 512$ patches. For datasets with original resolutions larger than $512 \times 512$ (\eg, WHU-CD\cite{ji2018fully}, HRSCD\cite{daudt2019multitask}), we partition them into non-overlapping $512 \times 512$ patches (\eg, intervals [0, 512], [512, 1024]), allowing overlap only for the boundary patches to ensure full coverage.

Specifically, the dataset splits are organized as follows: (1)~LEVIR-CD~\cite{chen2020spatial} focuses on building changes in urban/suburban areas at 0.5m resolution. After cropping to $512 \times 512$, the dataset contains 2,548 training and 1,392 testing image pairs. (2)~WHU-CD~\cite{ji2018fully} provides a single large-scale aerial image (32,507$\times$15,354) with extensive building construction/demolition annotations at 0.075m resolution. Following the official protocol, we partition it into $512 \times 512$ patches, yielding 1,260 training and 690 testing samples. (3)~DSIFN-CD~\cite{zhang2020deeply} covers multi-scale changes across diverse scenes at $\sim$0.03-1m resolution, emphasizing complex rural-urban transitions. The official split provides 3,600 training, 340 validation, and 48 testing image pairs, all at $512 \times 512$ resolution. (4)~SECOND~\cite{yang2021asymmetric} supports both BCD and SCD tasks with fine-grained semantic annotations for 7 land-cover categories. After standardizing to $512 \times 512$, we adopt the official split of 2,968 training and 1,694 testing pairs. (5)~HRSCD~\cite{daudt2019multitask} offers very-high-resolution (0.5\,m) imagery with dense pixel-wise change labels. The original 291 large images ($10{,}000 \times 10{,}000$ pixels) are first cropped into 116,400 patches of $512 \times 512$ pixels. HRSCD exhibits substantial class imbalance, with most labeled regions remaining unchanged across time; for example, changes involving artificial surfaces and agricultural land account for only 0.6\% of all change instances~\cite{zhao2022spatially}. Following~\cite{zhu2025change3d}, we therefore retain only image pairs in which changed areas occupy at least 20\% of the patch, resulting in a curated split of 905 training, 116 validation, and 123 testing samples. For zero-shot evaluation, models are trained exclusively on each synthetic dataset and directly evaluated on these real-world benchmarks without fine-tuning, ensuring a strict assessment of cross-domain generalization.

\subsection{Synthetic Datasets for Comparison}
To evaluate the downstream utility of KnowChange-generated data, we compare our synthetic datasets with representative change data synthesis datasets used for binary and semantic change detection. These datasets differ in their source data, synthesis strategy, scale, and supported task types. Table~\ref{tab:synthetic_datasets} summarizes the synthetic datasets involved in our experiments, including existing datasets and the three datasets generated by KnowChange, namely Know-BCD, Know-SEC, and Know-HR.
\begin{table*}[htbp]
\centering
\caption{Overview of synthetic change detection datasets. BCD and SCD denote binary and semantic change detection.}
\label{tab:synthetic_datasets}
\setlength{\tabcolsep}{3mm}
\resizebox{\textwidth}{!}{%
\begin{tabular}{llll}
\toprule
Synthetic Dataset & Base Dataset & \#Samples & Task \\
\midrule
WHU-GCD & LoveDA, Evlab-SS, LandCover.ai, Google Earth & 25K  &SCD \\
SyntheWorld & Fully Synthetic, OpenEarthMap & 40K & BCD \\
FSC-180k & FLAIR & 180K & SCD \\
Changen2-S1/S9 & xView2/OpenEarthMap & 15K/27K & BCD/SCD \\
Know-BCD/SEC/HR & OpenEarthMap, FLAIR & 10K/10K/10K & BCD/SCD/SCD \\
\bottomrule
\end{tabular}%
}
\end{table*}

\subsection{Layout-to-Mask Model Training}
The primary objective of training the Layout-to-Mask (L2M) model is to enable the acquisition of rich category-shape associations and ensure consistency between object shapes and their surrounding context. The model takes as input the original semantic map $S_{\text{pre}}$, a semantic mask $M_{\text{sa}}$ indicating regions to be inpainted, and textual prompts describing the target categories. To enhance the model's generalization capability across diverse scenarios, we design two complementary training strategies:

\textbf{Category-Aware Instance Masking:} We select specific semantic categories (either single or multiple) and identify their connected components. This strategy is implemented through two masking schemes:
    \begin{itemize}
        \item \textit{Bounding Box Masking:} Connected components are masked using their minimum bounding boxes. This forces the model to infer plausible shape distributions of specified categories within rectangular constraints.
        \item \textit{Connected Component Masking:} Connected components are masked using their exact shapes rather than bounding boxes. This focuses on fine-grained shape fidelity and precise boundary reconstruction.
    \end{itemize}
    For both schemes, the text prompt corresponds directly to the masked category names (\eg., ``Add building and road in the masked area''), enabling the model to learn specific category-shape priors with explicit semantic guidance.
    
\textbf{Random Region Masking:} We apply random rectangular masks that may span multiple semantic categories without prior category selection. In this case, the text prompt is defined as the dominant categories (top two by pixel count) within the masked region. This strategy encourages the model to maintain smooth spatial transitions and handle complex contextual relationships across different semantic boundaries without explicit category guidance.

For all strategies, we employ a unified text prompt template that specifies the color legend for semantic categories and instructs the model to generate content within the masked region. The key difference lies in how category names are specified in the prompt: For category-aware instance masking, the prompt includes the specific masked category names, providing explicit semantic guidance for shape generation. For random region masking, the prompt includes the top two dominant categories within the masked region, requiring the model to infer appropriate category placement based on context.

For example, a typical prompt follows the format:
\begin{quote}
    \small
    ``This is a remote sensing semantic map. The color legend is: building is red, road is gray, water is blue and grass is green. Add building and road in the masked area and use right color.''
\end{quote}

\begin{figure}[!t]
    \centering
    \includegraphics[width=\linewidth]{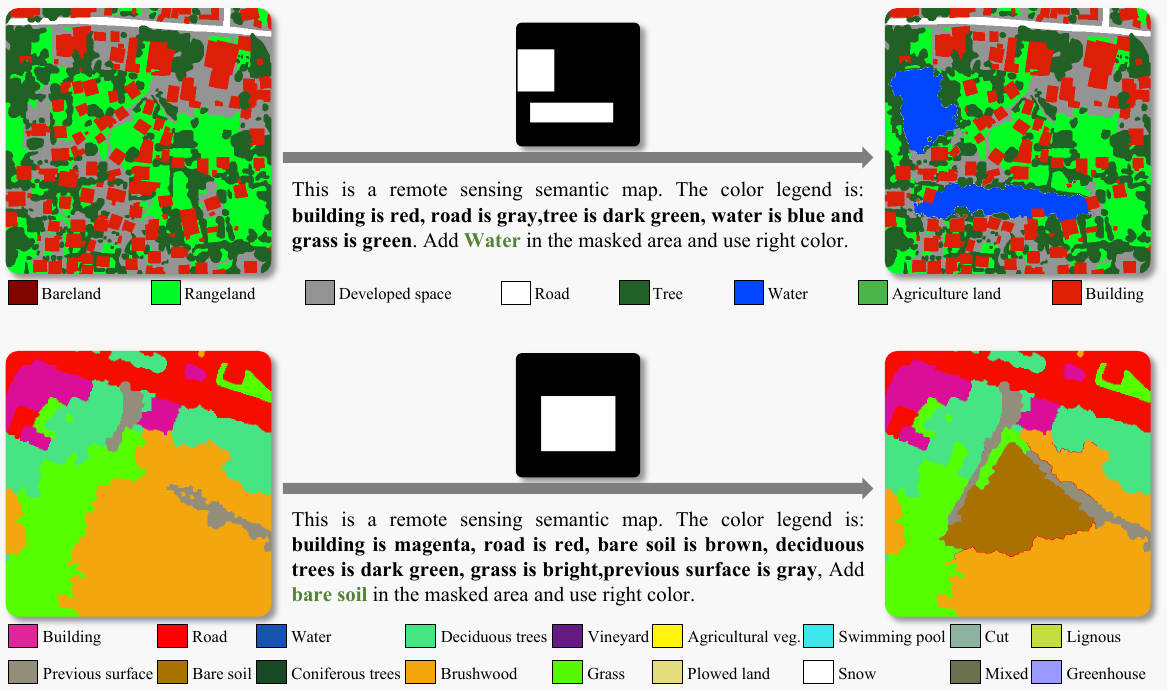}
    \caption{Examples of shape-altering transitions generated by the Layout-to-Mask model. The model fills masked regions with specified categories while maintaining shape plausibility and contextual consistency.}
    \label{fig:shape_alter_example}
\end{figure}

To prevent the model from memorizing fixed color-category bindings, we randomly assign colors to semantic categories across training samples. This design encourages the model to learn semantic category-shape associations through textual prompts rather than relying on color cues alone. Conditioned on the prompt, mask, and visible context, the model completes the masked regions. Having been trained on large-scale datasets, the model learns category-specific contour information and spatial distribution patterns. This enables the model to controllably fill specified categories within the target region $M_{\text{sa}}$ with plausible shapes, while maintaining consistency with surrounding objects. Specific examples illustrating this process are shown in Fig.~\ref{fig:shape_alter_example}.

\subsection{Mask-to-Image Model Training}

\begin{table*}[htbp]
\centering
\caption{Label mapping from the source datasets (OpenEarthMap and FLAIR) to the target categories of SECOND.}
\label{tab:label_mapping_second}
\setlength{\tabcolsep}{1mm}
\resizebox{\textwidth}{!}{%
\begin{tabular}{llll}
\toprule
OpenEarthMap Categories & FLAIR Categories & Target (SECOND) & Target ID \\
\midrule
Building & Building & Building & 5 \\
\cmidrule(lr){1-4}
Agriculture land, Rangeland & Agri. veg., Plowed land, Vineyard, Grass, Brushwood & Low vegetation & 1 \\
\cmidrule(lr){1-4}
Road, Developed space, Bareland & Road, Pervious, Bare soil, Cut, Mixed & Non-vegetated ground & 2 \\
\cmidrule(lr){1-4}
Tree & Coniferous, Deciduous trees, Lignous & Tree & 3 \\
\cmidrule(lr){1-4}
Water & Water, Swimming pool & Water & 4 \\
\cmidrule(lr){1-4}
Background & Greenhouse, Snow & Non-change & 0 \\
\bottomrule
\end{tabular}%
}
\end{table*}
 
Motivated by masked-image reconstruction strategies widely used in image editing~\cite{nichol2021glide,suvorov2022resolution}, we train the Mask-to-Image (M2I) model to reconstruct remote sensing images from masked inputs under semantic guidance. Similar reconstruction-based training schemes have also been adopted in remote sensing change synthesis, such as WHU-GCD~\cite{zan2025open} and HySCDG~\cite{benidir2025change}. Specifically, during training, the model takes a masked image $\widetilde{I}_{\mathrm{pre}}$, a re-rendering mask $M$, and the corresponding semantic mask $S_{\mathrm{pre}}$ as inputs, and reconstructs the original image $I_{\mathrm{pre}}$. This objective enables the model to synthesize category-consistent appearances within masked regions while preserving the visible surrounding context. During inference, $S_{\mathrm{pre}}$ is replaced with the synthesized post-change semantic mask $S_{\mathrm{post}}$ to generate $I_{\mathrm{post}}$. Following these established practices, we employ instance-, regional-, and global-level masking, together with semantic vector dropout, to cover diverse spatial contexts and semantic conditions.

\textbf{Instance-Level Masking:} We identify connected components in the semantic mask and randomly select multiple components during each training iteration. The corresponding image regions are masked using their exact component boundaries rather than bounding boxes, while their original semantic categories are retained as conditioning signals. The model reconstructs the masked image regions based on the surrounding image context and the pixel-wise semantic embeddings derived from the semantic mask. By exposing the model to objects with diverse shapes, sizes, and textures, this strategy facilitates category-specific appearance learning. Extremely small components are filtered out to reduce the influence of noisy artifacts.

\textbf{Regional-Level Masking:} We apply random rectangular masks that may cover multiple semantic categories. The size and aspect ratio of each mask are randomly sampled to provide diverse regional contexts. The model reconstructs the masked image region based on the visible surrounding context and the semantic embeddings within the masked region. This strategy encourages the model to generate coherent appearances across semantic boundaries, such as building--road and vegetation--water interfaces.

\textbf{Global-Level Masking:} The entire image is occasionally masked, leaving no visible image context. The model is therefore required to reconstruct the scene using only the semantic condition. This strategy strengthens holistic scene modeling and encourages the model to preserve spatial relationships among semantic categories, such as roads connected to buildings and water bodies adjacent to vegetation.

\textbf{Semantic Vector Dropout:} In addition to spatial masking, we apply semantic vector dropout to improve robustness to incomplete or noisy semantic conditions. During training, the semantic embeddings of randomly selected regions are replaced with zero vectors, requiring the model to infer plausible visual content from the visible image context and the semantic information of surrounding regions. This strategy reduces the model's sensitivity to missing or inaccurate semantic annotations.

For all training strategies, the semantic mask is converted into pixel-wise CLIP semantic embeddings, which are further transformed by the adapter described in the main paper into control features for the M2I model. The model is optimized to reconstruct the original image from the masked input under semantic conditioning. By combining instance-, regional-, and global-level masking with semantic vector dropout, M2I learns category-consistent appearances while preserving local boundary coherence and global contextual consistency.

\subsection{Training Details for Change Detection Models}
\textbf{Binary Change Detection Training:}
For BCD evaluation on datasets including LEVIR-CD, WHU-CD, DSIFN-CD, and SEC-BCD, we adopt ChangeFormer~\cite{bandara2022transformer} as the primary evaluation model. Since several synthetic datasets are derived from multi-category semantic segmentation sources, we extract binary change masks by considering only building-related transitions. Specifically, for our Know-BCD and Know-SEC, as well as other semantic-change-based synthetic datasets such as WHU-GCD~\cite{zan2025open}, FSC-180k~\cite{benidir2025change}, and Changen2-S9~\cite{zheng2024changen2}, a pixel is labeled as ``changed'' if its semantic category transitions involve building emergence, demolition, or expansion; all other semantic transitions are treated as unchanged. In contrast, datasets that are inherently building-focused, such as SyntheWorld~\cite{song2024syntheworld} and Changen2-S1~\cite{zheng2024changen2}, already provide binary change masks and require no additional processing. This unified protocol ensures consistency across all training data while leveraging the rich semantic information from source datasets.

\begin{table*}[htbp]
\centering
\caption{Label mapping from the source datasets (OpenEarthMap and FLAIR) to the target categories of HRSCD.}
\label{tab:label_mapping_hrscd}
\setlength{\tabcolsep}{1.6mm}
\resizebox{\textwidth}{!}{%
\begin{tabular}{llll}
\toprule
OpenEarthMap Categories & FLAIR Categories & Target (HRSCD) & Target ID \\
\midrule
Building, Road, Developed space & Building, Road, Pervious surface & Artificial surface & 1 \\
\cmidrule(lr){1-4}
\shortstack[l]{Agriculture land\\ Rangeland, Bareland}
&
\shortstack[l]{Agri. veg., Plowed land, Vineyard\\ Grass, Brushwood, Bare soil, Cut, Mixed}
&
Agricultural land
&
2 \\

\cmidrule(lr){1-4}
Tree & Coniferous, Deciduous trees, Lignous & Forest & 3 \\
\cmidrule(lr){1-4}
-- & Snow & Wetland & 4 \\
\cmidrule(lr){1-4}
Water & Water, Swimming pool & Water & 5 \\
\cmidrule(lr){1-4}
Background & Greenhouse, Unknown & No info & 0 \\
\bottomrule
\end{tabular}%
}
\end{table*}

\textbf{Semantic Change Detection Training:}
For SCD evaluation on SECOND and HRSCD datasets, we use Change3D~\cite{zhu2025change3d} for pixel-wise land-cover transition prediction. Since different datasets follow different class definitions, we apply label mapping rules to align the synthetic data with target benchmarks. The mapping tables from OpenEarthMap and FLAIR to SECOND and HRSCD categories are provided in Table~\ref{tab:label_mapping_second} and Table~\ref{tab:label_mapping_hrscd}, respectively. Specifically for the SECOND dataset, in addition to label mapping, we compare pre- and post-change semantic masks to identify unchanged regions, which are explicitly set to the non-change category to align with the benchmark protocol.

\section{Additional Experimental Results and Analysis}
\label{additional-results}
\subsection{Analysis of Change Proportions and Category Transition Matrices}
\begin{figure*}[!t]
  \centering
  \includegraphics[width=0.9\linewidth]{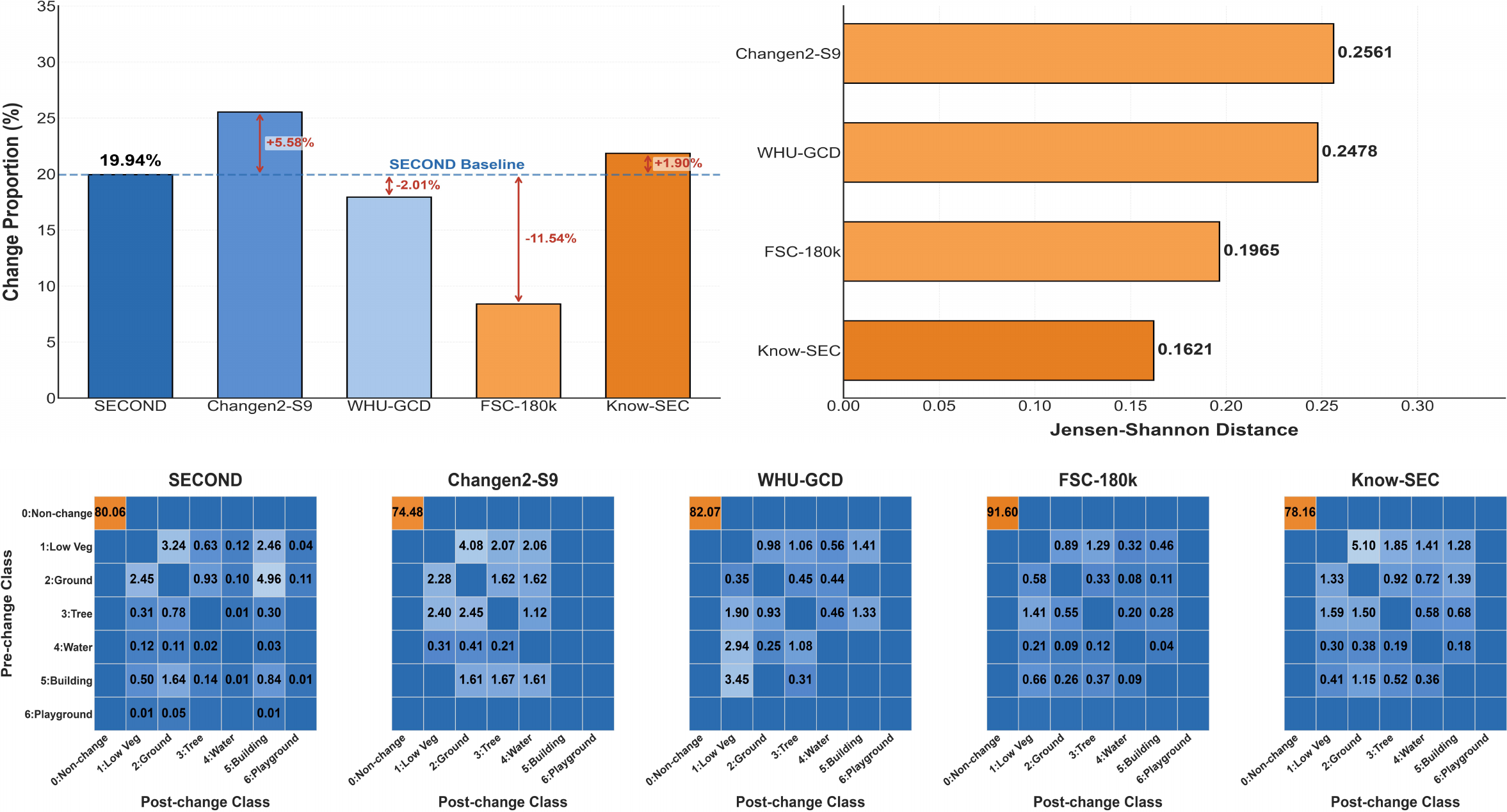}
  \caption{Overview of the statistical comparison between the real SECOND dataset and synthetic datasets regarding change proportions and category transition differences.}
  \label{fig:appendix_categorytransition}
\end{figure*}

To quantitatively evaluate the semantic diversity and realism of the synthesized change data, we conduct a comprehensive analysis involving category transition matrices, overall change proportions, and the Jensen-Shannon (JS) divergence. Since previous studies have designed specific rules for category transitions\cite{zheng2024changen2,zan2025open}, this analysis aims to compare the differences in category changes between various synthetic datasets and the real dataset. All datasets are mapped to the unified SECOND taxonomy for a fair comparison. The visualization results comparing the real SECOND dataset with four synthetic datasets (Changen2-S9, WHU-GCD, FSC-180k, and our Know-SEC) are shown in Fig.~\ref{fig:appendix_categorytransition}.

The SECOND dataset serves as the real-world baseline with a change proportion of 19.94\%. Among the synthetic datasets, Changen2-S9 exhibits the highest proportion at 25.52\%, whereas FSC-180k records the lowest at approximately 8.4\%. Both WHU-GCD and our Know-SEC align most closely with the baseline, showing marginal differences of 2.01\% and 1.90\%, respectively. This proximity suggests that our synthesized data closely approximates the change density found in real-world scenarios. Additionally, regarding distributional similarity, our Know-SEC achieves the lowest JS distance of 0.1621 compared to the SECOND dataset.

The class transition matrices further reveal clear differences in transition coverage among the synthetic datasets. SECOND contains diverse inter-class and intra-class transitions, reflecting the complexity of real-world changes. Although Changen2-S9 and WHU-GCD achieve change proportions of 25.52\% and 17.93\%, respectively, their transition matrices remain relatively sparse, with several class transitions observed in SECOND being absent. This is mainly because their change simulation relies on predefined transition matrices or handcrafted rules under fixed category settings. Consequently, transitions not explicitly covered by these designs cannot be adequately synthesized, reflecting the limited transition coverage and inflexibility discussed in the main paper. FSC-180k covers a broader range of inter-class transitions, but its change proportion is only 8.40\%, substantially lower than the 19.94\% observed in SECOND. Its transition frequencies are also highly imbalanced, indicating that expanding the predefined transition set alone does not ensure a realistic transition distribution.

In contrast, Know-SEC does not restrict change simulation to a predefined transition matrix. Instead, the VLM infers plausible class transitions from the scene context and textual prompt, resulting in broader transition coverage and a change proportion of 21.84\%, which differs from SECOND by only 1.90 percentage points. Know-SEC consequently achieves the lowest JS distance of 0.1621 among the compared synthetic datasets, indicating the closest overall transition distribution to the real dataset. Nevertheless, some distributional imbalance remains. For example, the Low Vegetation-to-Ground transition accounts for 5.10\% in Know-SEC, compared with 3.24\% in SECOND, and its intra-class transitions remain less diverse. This discrepancy may arise because KnowChange focuses on the contextual plausibility of individual changes rather than explicitly matching empirical transition frequencies, while the class distribution of the source semantic data also affects the generated distribution.

\subsection{Effectiveness of In-Domain Data Augmentation:}
\begin{figure}[!t]
    \centering
    \includegraphics[width=0.85\linewidth]{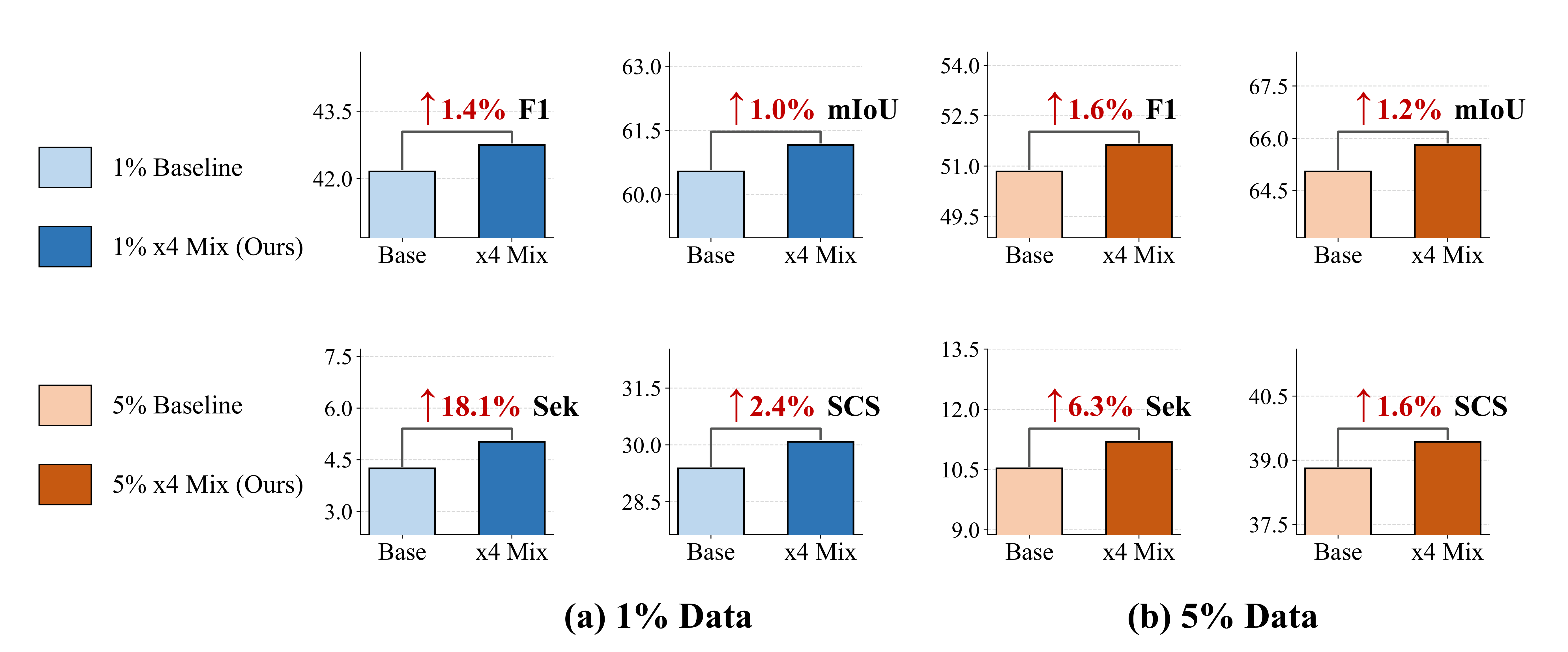} 
\caption{Impact of 4$\times$ in-domain augmentation under different labeled data settings.}
    \label{fig:second_indomain}
\end{figure}

Different from methods such as Changen2 and HySCDG, KnowChange leverages VLMs as its knowledge source and is trained over a broad semantic vocabulary, enabling it to perform in-domain augmentation on previously unseen datasets. To evaluate this capability, we randomly sample 1\% and 5\% of the labeled training samples from SECOND, a dataset not seen during the training of KnowChange. For each setting, KnowChange uses the sampled images and their labels to synthesize target-domain change samples at four times the size of the corresponding subset. The generated samples are then combined with the sampled real data to train the downstream model. As shown in Fig.~\ref{fig:second_indomain}, incorporating the synthesized data improves all four metrics under both labeled-data settings. SeK exhibits the most pronounced gains, increasing by 18.1\% and 6.3\% under the 1\% and 5\% settings, respectively. These results suggest that few-shot in-domain augmentation with KnowChange can partially alleviate category imbalance and limited supervision on unseen datasets, resulting in modest improvements in semantic change detection performance.

\subsection{Generalization to Additional Remote Sensing Tasks}
\begin{figure}[!t]
  \centering
  \includegraphics[width=0.9\linewidth]{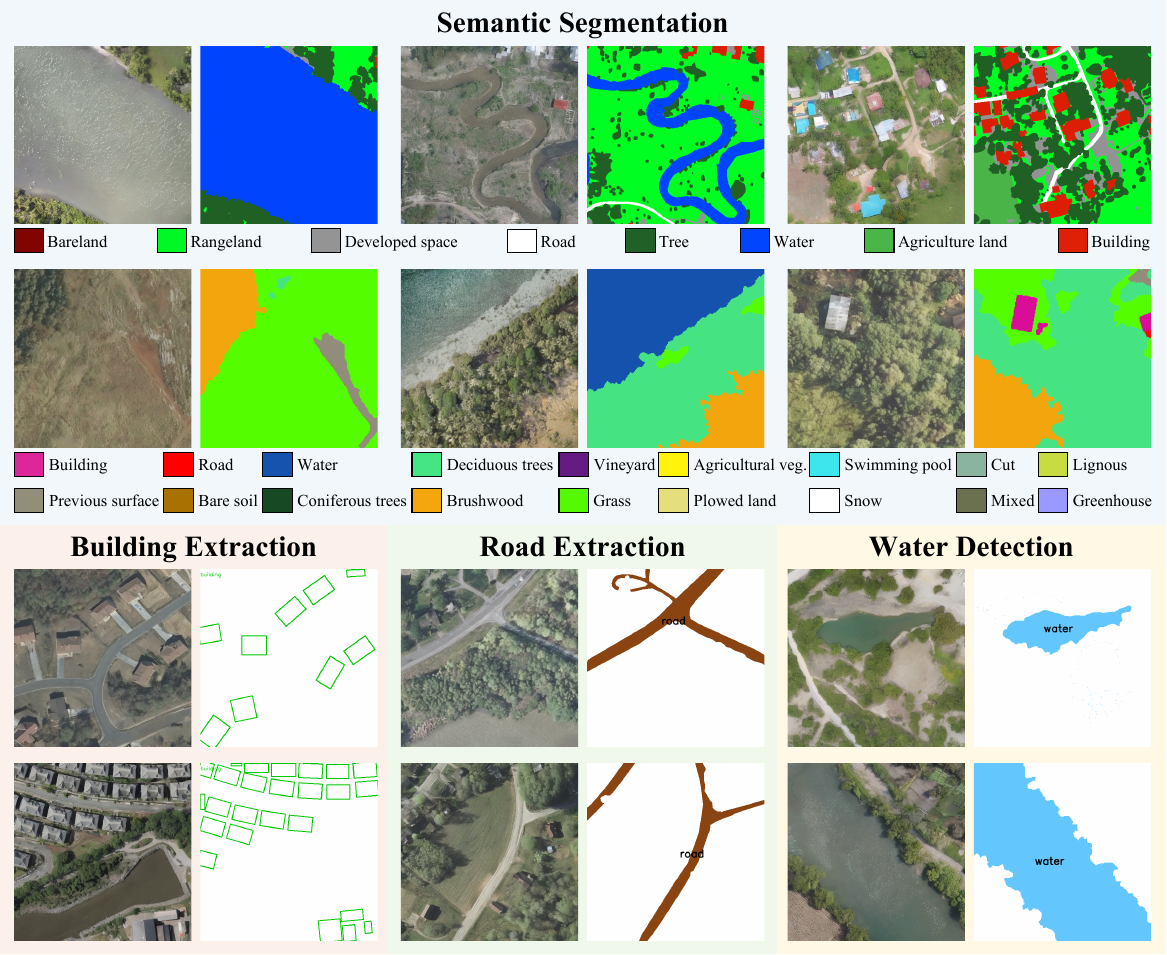}
  \caption{Visualization of multi-task capabilities using our framework. These samples are generated directly from semantic maps without pre-temporal image inputs ($I_{\text{pre}}$). The visualization demonstrates applicability to downstream tasks including semantic segmentation, building extraction, road extraction, and water detection.}
  \label{fig:multi_task}
\end{figure}

The main experiments focus on building and semantic change detection, leaving the applicability of the generalizable semantic-guided synthesis component beyond change data construction less explored. To complement these evaluations, we further examine whether KnowChange can support other remote sensing tasks through semantic-map-conditioned image synthesis. Without requiring real images as references, the framework can generate remote sensing images from semantic maps while preserving pixel-level correspondence between the generated images and input annotations. This property enables the direct construction of image--annotation pairs for semantic segmentation and category-specific land-cover extraction, such as buildings, roads, and water bodies. As illustrated in Fig.~\ref{fig:multi_task}, KnowChange produces visually plausible images that remain consistent with the provided semantic layouts across these tasks. These qualitative results complement the main experiments by demonstrating that the proposed semantic-guided synthesis framework is not restricted to change detection and can potentially support data construction for a broader range of remote sensing tasks.

\subsection{Image Reconstruction Quality Analysis}
The M2I model is responsible for rendering post-change images according to semantic masks while preserving the content of unchanged regions. To evaluate its image synthesis capability, we use M2I, HySCDG, and Changen2 to reconstruct images from the SECOND dataset under the same evaluation setting. The reconstructed images are compared with their corresponding real images using PSNR, SSIM, and LPIPS.

\begin{table}[h]
\centering
\caption{Image reconstruction quality comparison on the SECOND dataset.}
\label{tab:M2I_comparison}
\setlength{\tabcolsep}{2mm}
\begin{tabular}{lccc}
\toprule
Model & PSNR $\uparrow$ & SSIM $\uparrow$ & LPIPS $\downarrow$ \\
\midrule
M2I & \textbf{22.3340} & \textbf{0.7042} & \textbf{0.1877} \\
HySCDG & 21.3064 & 0.6900 & 0.1960 \\
Changen2 & 22.1237 & 0.6985 & 0.1968 \\
\bottomrule
\end{tabular}
\end{table}

As shown in Tab.~\ref{tab:M2I_comparison}, M2I consistently outperforms HySCDG and Changen2 across all three metrics. Compared with the strongest competing result for each metric, M2I improves PSNR and SSIM by 0.2103 and 0.0057, respectively, while reducing LPIPS by 0.0083. The higher PSNR and SSIM indicate better preservation of image content and spatial structure, whereas the lower LPIPS reflects improved perceptual similarity to the real images. These results demonstrate that M2I can more accurately reconstruct remote sensing images from semantic conditions, supporting its ability to generate structurally consistent and visually realistic change data.

\subsection{Long-Term Urban Evolution Synthesis}
\begin{figure*}[!t]
    \centering
    \includegraphics[width=0.94\textwidth,height=0.62\textheight,keepaspectratio]{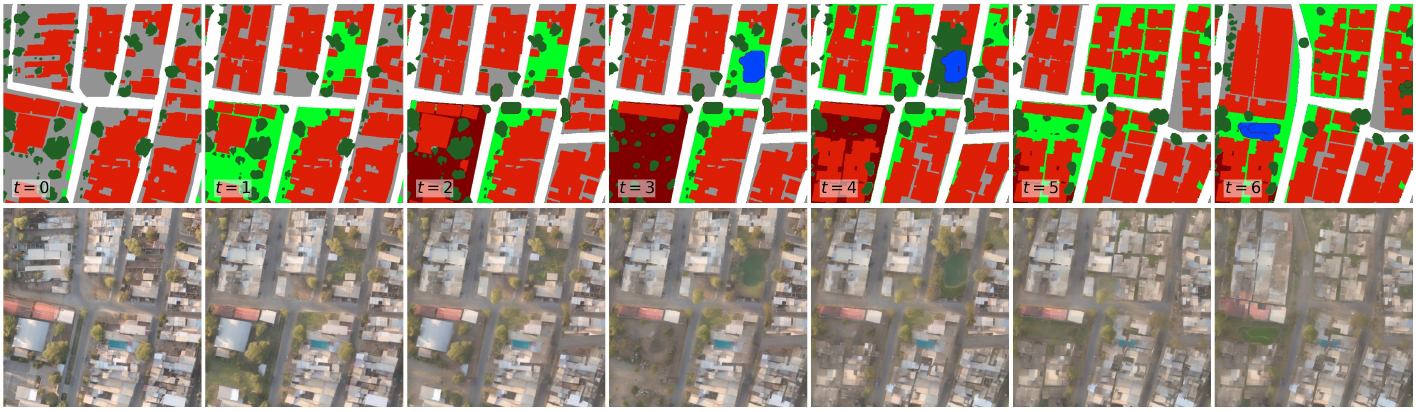}
    \caption{Long-term urban evolution synthesized from a single-temporal scene. KnowChange iteratively generates semantic maps (top) and their corresponding remote sensing images (bottom), producing multi-category changes while maintaining semantic and spatial consistency across time.}
    \label{fig:long_termV1_supp}
\end{figure*}

Beyond bi-temporal change data synthesis, KnowChange can be iteratively applied to transform a single-temporal scene into a long-term multi-temporal sequence. Specifically, the image and semantic map generated at each timestamp are used as the input for the subsequent timestamp, allowing the scene to evolve progressively over time. As shown in Fig.~\ref{fig:long_termV1_supp}, the synthesized sequence presents complex urban evolution involving multiple land-cover categories, including buildings, vegetation, trees, roads, and water bodies. Across successive timestamps, the spatial distribution of these categories changes gradually, while the overall road structure and unchanged regions remain coherent. The generated images also remain well aligned with their corresponding semantic maps throughout the sequence. These results demonstrate the capability of KnowChange to model long-horizon, multi-category urban evolution while preserving temporal, spatial, and semantic consistency, extending its applicability from bi-temporal change synthesis to multi-temporal change simulation.

\section{Related Work}
\label{related-work}

\textbf{Binary \& Semantic Change Detection:}
Change detection in remote sensing imagery identifies differences across temporal observations, evolving from traditional algebraic methods to deep learning-based approaches. Binary change detection (BCD) methods initially adopted Siamese CNNs~\cite{daudt2019multitask,zhan2017change,daudt2018fully} and were subsequently enhanced with attention mechanisms~\cite{fang2021snunet,chen2020spatial} and Transformers~\cite{chen2021remote,bandara2022transformer}. These methods focus on locating changed regions but do not characterize the associated semantic transitions. Semantic change detection (SCD) extends BCD by jointly identifying change locations and land-cover categories~\cite{daudt2019multitask,yang2021asymmetric}. However, SCD remains challenged by severe class imbalance, scarce annotations for rare categories, and the complex transition space between land-cover classes~\cite{TIAN2022164,zan2025open}. Transfer learning partially alleviates annotation scarcity by exploiting models pretrained on large-scale source domains~\cite{cao2023full,TIAN2022164}, but both BCD and SCD still rely heavily on pixel-level annotations. In particular, rare and diverse change transitions are difficult to cover with real training samples, motivating change data generation as a scalable means of expanding training distributions~\cite{zang2025changediff,benidir2025change,zan2025open,zheng2024changen2}.

\textbf{Data Generation for Change Detection:}
Data generation alleviates the scarcity of annotated datasets. Fully synthetic methods use 3D rendering engines such as SyntheWorld~\cite{song2024syntheworld} to create scenes from scratch, offering explicit parameter control but often lacking real-world texture diversity. Hybrid approaches instead insert synthetic changes into real images. Early methods employed copy-paste operations, such as ChangeStar~\cite{zheng2021change} and Self-Pair~\cite{seo2023self}, which may introduce visible artifacts and inconsistent object-context relationships. IAug~\cite{chen2021adversarial} utilizes GANs to simulate building changes but has limited applicability to diverse land-cover transitions. Recent approaches leverage diffusion models to improve synthesis quality. Changen~\cite{zheng2023changen} and Changen2~\cite{zheng2024changen2} synthesize bi-temporal images conditioned on semantic layouts, while FSC-180k~\cite{benidir2025change} combines Stable Diffusion with ControlNet~\cite{zhang2023adding} for semantic-guided inpainting. SD-Inpainting-RS~\cite{zan2025open} further employs predefined semantic rules and geometric constraints to guide change generation. Text-guided methods such as RSDiff~\cite{sebaq2024rsdiff} and DiffusionSat~\cite{khanna2023diffusionsat}, as well as single-temporal generators such as AeroGen~\cite{tang2025aerogen} and TerraGen~\cite{tang2025terragen}, can synthesize high-quality remote sensing images but are not designed to construct spatially aligned bi-temporal change pairs. Despite these advances, existing change synthesis methods commonly rely on handcrafted transition rules with limited coverage and require repeated customization for different change types. KnowChange instead leverages pretrained VLMs as knowledge sources to infer plausible change locations and class transitions from scene contexts and desired change types. Combined with generalizable semantic-guided synthesis models, this design enables flexible generation of diverse change data without repeatedly redesigning the synthesis pipeline.

\end{document}